\documentclass[final]{l4dc2024}

\title[\textit{Proto}-MPC]{\textit{Proto}-MPC: An Encoder-Prototype-Decoder Approach for Quadrotor Control in Challenging Winds}

\usepackage{times}
\usepackage{wrapfig}
\usepackage{algorithm,algpseudocode}
\usepackage{bbm}
\usepackage{multirow}
\usepackage{multicol}
\usepackage{caption}
\usepackage{multirow,makecell}
\usepackage{booktabs}

\newcommand{\norm}[1]{\lVert #1 \rVert}

\coltauthor{\Name{Yuliang Gu} \Email{yuliang3@illinois.edu}\\
 \Name{Sheng Cheng} \Email{chengs@illinois.edu}\\
 \Name{Naira Hovakimyan} \Email{nhovakim@illinois.edu}\\
 \addr Mechanical Science and Engineering, University of Illinois Urbana-Champaign, Urbana, IL 61801}

\begin{document}
\maketitle

\begin{abstract}%
Quadrotors are increasingly used in the evolving field of aerial robotics for their agility and mechanical simplicity.
However, inherent uncertainties, such as aerodynamic effects coupled with quadrotors' operation in dynamically changing environments, pose significant challenges for traditional, nominal model-based control designs. To address these challenges, we propose a multi-task meta-learning method called Encoder-Prototype-Decoder (EPD), which has the advantage of effectively balancing shared and distinctive representations across diverse training tasks.
Subsequently, we integrate the EPD model into a model predictive control problem (\textit{Proto}-MPC) to enhance the quadrotor's ability to adapt and operate across a spectrum of dynamically changing tasks with an efficient online implementation. 
We validate the proposed method in simulations, which demonstrates \textit{Proto}-MPC's robust performance in trajectory tracking of a quadrotor being subject to static and spatially varying side winds.
\end{abstract}


\section{Introduction}
In the evolving field of aerial robotics, quadrotors are widely used due to their agility and versatility in various applications. To fully leverage the agility of quadrotors, controller designs are heavily based on quadrotor models. Generally, these models are derived following the Newton-Euler equations, which can hardly accommodate dynamic uncertainties in real-world applications (e.g., wind, aerodynamic effects, slung or slosh payloads). To address this limitation, recent research has focused on using advanced machine learning methods, such as Gaussian Process~\citep{torrente2021data} and NeuralODE~\citep{chee2022knode}, to learn an accurate dynamical model from real-world data and integrate it with model-based control design, which can significantly enhance the system performance.

Quadrotors operating in real-world scenarios frequently encounter a range of \textit{structurally similar yet appearingly different} tasks, each with unique dynamical uncertainties. For instance, a quadrotor might face varying side wind conditions or be tasked with transporting slung payloads of unknown mass. These varied tasks pose a unique challenge for the above-mentioned control methods. While relying on a single data-driven model often falls short of achieving optimal performance across diverse scenarios, training multiple models for case-specific tasks is inefficient due to 1) challenges in data collection for each specific case and 2) potentially time-consuming online switches of different trained models that use a relatively large amount of parameters for each individual task.
To tackle these challenges, a growing line of research investigates the use of online learning and meta-learning techniques. These methods operate in an offline-online framework~\citep{o2022neural,jiahao2023online, richards2021adaptive, wang2024neural}, allowing for adaptation of the learned models or real-time retraining of new models to align with the changing characteristics of operational tasks. (A more detailed literature review is available in the Appendix.)

Integrating online learning methods into model-based control design poses several key challenges: 1) \textbf{adaptivity}: the system must rapidly respond to real-time changing conditions; 2) \textbf{model fidelity}: as data-driven models evolve through online learning, they risk losing essential knowledge learned from the initial training data, which can lead to unpredictable behaviors and reduced performance in situations that they were originally designed to handle; 3) \textbf{exploration vs. exploitation}: reaching the right balance between exploring new data and exploiting existing knowledge is critical to ensure reliable real-time performance.

\begin{figure}
    \centering
\includegraphics[width=1.0\textwidth]{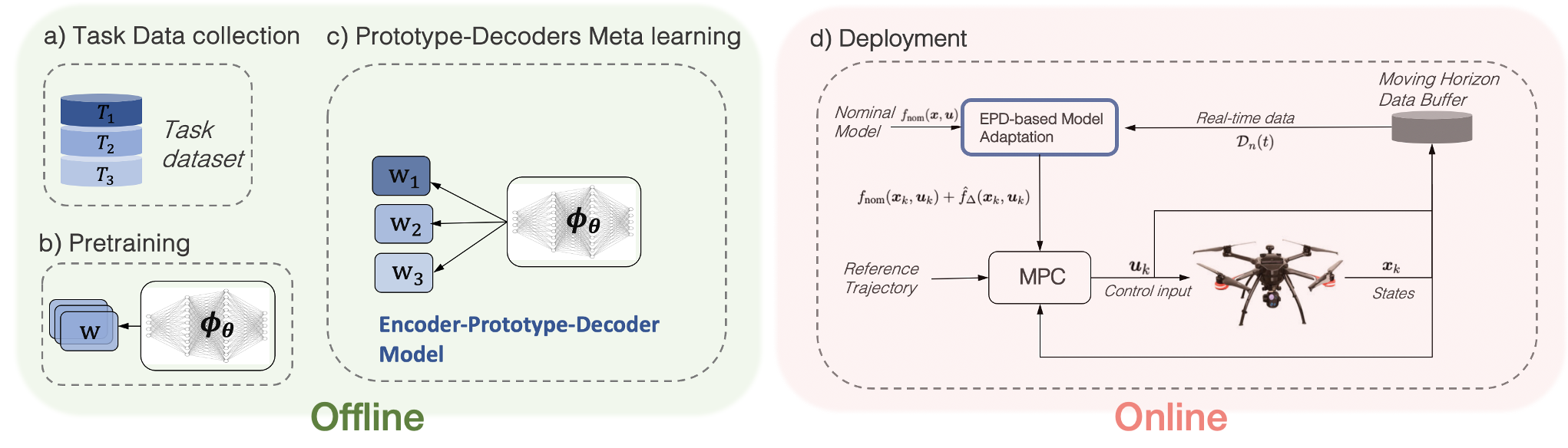}
    \caption{Framework Overview. a) Collecting data on multiple tasks; b) Pretraining to ensure that encoder-decoder pairs can capture the overall patterns of the data; c) Jointly training task-specific prototype decoders to capture distinctive task features and regularizing the encoder to avoid overfitting; d) Online implementation of Proto-MPC with prototype-decoder-based adaptation.}
    \label{fig: Framework overview}
    \vspace{-0.3 cm}
\end{figure}

To address these challenges, we introduce \textit{Proto}-MPC, a novel multi-task meta-learning-based model predictive control (MPC) framework. Central to our method is an \textbf{Encoder-Prototype-Decoder} (EPD) model, which is designed to learn the residual dynamics of the quadrotor from diverse tasks. The EPD model comprises two key components: a universal deep neural network (DNN) encoder and a set of task-specific linear prototype decoders.  On the one hand, the encoder learns the common and essential patterns across various task datasets, providing a generalized understanding of the tasks by producing their representations on a low-dimensional manifold (i.e., features). On the other hand, the linear prototype decoder captures the distinctive characteristics of a specific task in a computationally efficient way (due to its linear form). In the online inference stage, the encoder processes incoming data into features, while prototype decoders are used as a ``basis'' to interpolate encoded features as residuals in the dynamics. This architecture allows fast computation of a new decoder aligned with the current task's characteristics online. Moreover, this adaptive approach ensures the MPC has an accurate, up-to-date residual dynamical model. 
We evaluate the proposed framework on a quadrotor under various speeds of side wind. The results showcase the generalization and fast adaptation of the proposed \textit{Proto}-MPC framework.

The contributions of this paper are summarized as follows: we propose \textit{Proto}-MPC, a novel
model predictive control framework for quadrotor control subject to uncertainties and disturbances. We propose an
EPD model as a data-driven augmentation to the physics-based dynamics to capture the uncertainties. The EPD model can achieve the balance between generalizing across a wide array of tasks, both trained and unseen tasks, and rapidly adapting to dynamically evolving task conditions with tunable parameters.

\section{Background: Nonlinear MPC for Quadrotor Control}
We consider the 6 DoF rigid body dynamics of the quadrotor (with mass $m$ and inertia $J$)
\begin{equation}\label{eq: nominal dynamics of quadrotors}
    \dot{\boldsymbol{p}} = \boldsymbol{v}, \ \dot{\boldsymbol{v}} = m^{-1} f \boldsymbol{z}_B  + \boldsymbol{g} ,  \ \dot{\boldsymbol{q}} = \frac{1}{2} \boldsymbol{q} \otimes [
            0 \ \boldsymbol{\omega}^\top]^\top , \: \dot{\boldsymbol{\omega}} = 
J^{-1} (\boldsymbol{M} - \boldsymbol{\omega} \times J \boldsymbol{\omega}),
\end{equation}
where $\boldsymbol{p} \in \mathbb{R}^3$, $\boldsymbol{v} \in \mathbb{R}^3$ stand for the position and velocity of the quadrotor in the inertial frame, $\boldsymbol{q}=[q_0, q^{\top}]^{\top} \in \mathbb{S}^3$ (where $q_0 \in \mathbb{R}$ and $q\in \mathbb{R}^3$) is the unit quaternion for rotation from the inertial to body frame, and $\boldsymbol{\omega} \in \mathbb{R}^3$ is the angular velocity in body frame. 
The gravitational acceleration is denoted by $\boldsymbol{g}$.
The vector $\boldsymbol{z}_B$ is the unit vector aligning with the $z$-axis of the body frame. The state $\boldsymbol{x} = [\boldsymbol{p}^\top \ \boldsymbol{v}^\top \ \boldsymbol{q}^\top \ \boldsymbol{\omega}^\top]^\top$ follows a discretized version of the dynamics in~\eqref{eq: nominal dynamics of quadrotors} as ${\boldsymbol{x}}_{k+1} = f_{\text{nom}}(\boldsymbol{x_{k}}, \boldsymbol{u}_{k})$ with control being $\boldsymbol{u} = [f \ \boldsymbol{M}^\top]^\top \in \mathbb{R}^4$ (total thrust $f$ and moment $\boldsymbol{M} \in \mathbb{R}^3$). 

The model in \eqref{eq: nominal dynamics of quadrotors} describes the nominal case with no dynamical uncertainty. In general, uncertainties (e.g., wind or aerodynamic effects) exist in a real system. We consider lumped uncertainties (see~\citep{wu20221}), denoted by $f_{\Delta}$, in the dynamics to account for the impact to the system, resulting in the real dynamics $f_{\text{real}} = f_{\text{nom}} + f_{\Delta}$.
In this paper, we will learn the lumped uncertainties as $\hat{f}_{\Delta}$. The objective is to ensure that the learned dynamics $f_{\text{nom}} + \hat{f}_{\Delta}$ closely approximate the actual dynamics $f_{\text{real}}$, which allows us to use it as a trustworthy model in an MPC formulation. We consider the following nonlinear MPC
\begin{equation}
\begin{aligned}
    \boldsymbol{u}_{0:N-1}^\star =\ & \underset{\boldsymbol{u}_{0:N-1}}{\text{argmin}} && \sum_{k=0}^{N-1} \norm{\boldsymbol{x}_k - \bar{\boldsymbol{x}}_k}^2_Q + \norm{\boldsymbol{u}_k - \bar{\boldsymbol{u}}_k}^2_R + \norm{\boldsymbol{x}_N - \bar{\boldsymbol{x}}_N}^2_{Q_N} \\
    & \text{subject to} && \boldsymbol{x}_{k+1} = f_{\text{nom}}(\boldsymbol{x}_k,\boldsymbol{u}_k) + \hat{f}_{\Delta}(\boldsymbol{x}_k,\boldsymbol{u}_k), \ \boldsymbol{x}_0=\boldsymbol{x}_{\text{init}}, \\
    & && \boldsymbol{u}_{\min} \leq \boldsymbol{u} \leq \boldsymbol{u}_{\max},
\end{aligned}
\label{prob: MPC problem}
\end{equation}
where $\bar{\boldsymbol{x}}_k$ and $\bar{\boldsymbol{u}}_k$ denote the reference state and control, $Q$ and $R$ are the penalty matrices for deviating from the references, and $\boldsymbol{u}_{\min}$ and $\boldsymbol{u}_{\max}$ represent the limits on the control actions. 

\section{Method}
\subsection{Dataset}
Consider a set of $N$ tasks, $\mathcal{T} = \{T_k\}_{k=1:N}$. We are given their corresponding datasets, $\mathcal{D} = \{D^{T_k} \}_{k=1:N}$, where $D^{T_k} = \{ (x,y) \}^{T_k}$ consists of task-specific \textit{identically independently distributed} input-output pairs. The joint distribution of the input-output pairs $D^{T_k}$ is $P^{T_k}(x,y)$.
The task-specific batch data (of size $n$) is $D_n^{T_k}$, which is \textit{uniformly} sampled from $D^{T_k}$, denoted as $D_n^{T_k} \sim D^{T_k}$, and its empirical distribution is $P_n^{T_k}(x,y)$.

\subsection{Prototype-Decoder-Based Meta-Learning}
In our approach, we decompose the learned residual dynamics into the following form:
\begin{equation}\label{eq:residual_dyn}
y = \hat{f}_{\Delta}(\boldsymbol{x},\boldsymbol{u}) = w\phi_{\theta}(x),
\end{equation}
where $x = concat[\boldsymbol{x}, \boldsymbol{u}]$ represents the concatenated state and control input vectors, $\phi_{\theta}$ is an encoder, and $w$ is a linear decoder. Here, the $\phi_{\theta}$: $\mathbb{R}^{17} \rightarrow \mathbb{R}^p $ is a DNN parameterized by $\theta$ that encodes input data into a feature space in $\mathbb{R}^p$. The decoder
$w$ is a matrix with appropriate dimension and $w\in \mathcal{W} = \{w: \|w\|_2 = \sigma_{\max}(w) < w_0 \}$. The decoder maps the encoded features to the output as residuals in the dynamics.


We use the encoder-decoder as shown in \eqref{eq:residual_dyn} to capture the residual dynamics when a quadrotor conducts \textit{structurally similar yet appearingly different} tasks, such as flying in side-wind of different speeds. However, for an encoder-decoder pair with fixed parameters to adapt to different tasks, significant modifications or separate models may be required.
To tackle this multi-task scenario, we introduce the EPD model that comprises a task-agnostic encoder $\phi_{\theta}$ and a set of task-specific prototype encoders $\mathbf{W} = \{\mathbf{w}_k\}_{k=1:N}$. (Note that we use the bold font $\mathbf{w}$ to denote the prototype decoder, which should be distinguished from an arbitrary decoder denoted by $w$.) 
On the one hand, 
the encoder $\phi_{\theta}$ is trained to be task-agnostic in the sense that it captures the essential characteristics of all task datasets and allows for fast adjustments of the decoder.  
On the other hand, each prototype decoder $\mathbf{w}_k$ takes the encoded features and outputs precise task-relevant residuals, which essentially fine-tunes the EPD model to operate on the given task $T_k$. As key components in our method, prototype decoders are used as a ``basis'' to span a subspace in the task space, which enables 1) offline inter-task regularization and 2) online inter-task interpolation.



\subsection{Prototype Decoder}
In this subsection, we formally define and derive a prototype decoder. In brief, given an encoder $\phi_{\theta}$, a prototype decoder is the most representative of the given task data in some set of decoders. The representativeness of an encoder-decoder pair $(\phi_{\theta}, w)$ for a task $T_k$ is measured by its empirical risk on task $T_k$'s batch data:
\begin{equation}\label{eq:emp_risk}
    \mathcal{R}^{T_k}_{n}(w, \phi_{\theta}) = \frac{1}{n}\sum_{(x_i, y_i) \in D_n^{T_k} }\|y_i- \hat{y}_i \|^2,
\end{equation}
where $\hat{y}_i =w \phi_{\theta}(x_i)$, and $D_n^{T_k}$ is sampled from $D^{T_k}$. To ensure that the pair $(\phi_{\theta}, w)$ captures the overall data patterns effectively, the empirical risk must be bounded in a predefined threshold. We define this property as the \textit{achievability} condition as follows:
\begin{definition}\label{achi_set}
\textbf{(Achievability)} For a task $T_k \in \mathcal{T}$, an encoder-decoder pair $(\phi_{\theta},w)$ is achievable with some $R_0 \in \mathbb{R}^+$ if:
\begin{equation}\label{eq: achievability}
     \lim_{n\to \infty} \mathcal{R}_n^{T_k}(w, \phi_{\theta}) = \lim_{n\to \infty} \mathbb{E}_{D_n^{T_k}\sim D^{T_k}} \big [\mathcal{R}(w, \phi_{\theta})\big] = \lim_{n\to \infty} \int \mathcal{R}(w, \phi_{\theta}) dP_n^{T_k}(x,y) \leq R_0.
\end{equation}
\end{definition}
The achievability condition essentially imposes an upper bound on the expected risk to ensure that an encoder-decoder pair has a bounded error over the entire task dataset. One can pretrain the encoder by the alternating minimization method (minimization of the empirical risk by alternating between $\phi_{\theta}$ and $w$. More discussions are given in Remark~\ref{remark: information theory}) to satisfy the achievability condition. We summarize the pretraining procedure in Algorithm~\ref{training} in the Appendix. The pretraining step is critical in the sense that the model can learn from data in a ``lossy'' way while staying anchored to the core features.

Given an encoder $\phi_{\theta}$, we define the set of decoders satisfying~\eqref{eq: achievability} as a \textbf{task-achievable decoder set} $\mathcal{A}_{\phi_{\theta}}(T_k) = \big\{w \in \mathcal{W}: \lim_{n\to \infty} \mathcal{R}_n^{T_k}(w, \phi_{\theta}) < R_0\big\} $. 
The task-achievable decoder set $\mathcal{A}_{\phi_{\theta}}(T_k)$ consequently specifies a task-specific achievable region in $\mathcal{W}$. We are now ready to introduce a novel component called the \textit{\textbf{prototype decoder}}, which is a representative in $\mathcal{A}_{\phi_{\theta}}(T_k)$. The prototype decoder is a critical part of our model, aimed at effectively capturing the individual characteristics of each task:
\begin{definition}\label{proto}
\textbf{(Prototype Decoder)} For a given task $T^{k}\in \mathcal{T}$, the prototype decoder, denoted by $\mathbf{w}_k$, achieves the minimal empirical risk over the achievable set: $\mathbf{w}_k = \operatorname{argmin}_{w \in \mathcal{A}_{\phi_{\theta}}(T_k)} \mathcal{R}_n^{T_k}(w, \phi_{\theta})$.
\end{definition}
The prototype decoder captures the central characteristics of its corresponding task to achieve minimal risk among all the achievable decoders. This choice aligns with the principle of risk minimization, focusing on achieving the most efficient and effective learning outcome for each task. In practice, the prototype decoder can be computed empirically via
\begin{align}\label{eq:emp prototype}
    \mathbf{w}_{k,emp} = \operatorname{argmin}_{w \in \bar{\mathcal{W}}} \mathcal{R}^{T_k}_n(w, \phi_{\theta}),
\end{align}
where $\bar{\mathcal{W}}$ is finite set of achievable decoders. This empirical computation results in a geometric interpretation of the role of the prototype decoder: it is the geometric center of the achievable decoders under the ``distance'' defined by the risk, which is a concept that closely relates to Prototypical Networks~\citep{snell2017prototypical} for few-shots classification. Similarly the prototype decoder acts as a representative of the associated task in our EPD model framework.


\begin{remark}\label{remark: information theory}
In Rate-Distortion Theory, the definition of empirical risk in~\eqref{eq:emp_risk} is in fact a distortion measure between sequences~\citep{cover1999elements}. In our formulation, an achievable decoder set $\mathcal{A}_{\phi_{\theta}}(T_k)$ with an encoder $\phi_{\theta}$ specifies a rate-distortion region for a given task $T_k$. Moreover, the encoder-prototype-decoder pair ~\eqref{proto} is the rate-distortion function that achieves the infimum rate for a given distortion threshold $R_0$. The Blahut-Arimoto algorithm~\citep{arimoto1972algorithm} was proposed for calculating the rate-distortion function, which is an alternating minimization procedure. This algorithm can be specialized in our setting to pretrain the model to ensure achievability by alternating between encoder and decoder to minimize the empirical risk. In addition, such an achievability constraint in effect imposes an information bottleneck~\citep{tishby2000information} to balance the compression-representation trade-off. 
\end{remark}
With the prototype decoder effectively capturing task-specific characteristics, we next introduce a \textit{Prototype-Decoder Based Meta-Update} method to fine-tune the encoder. This approach prevents overfitting on the training tasks, ensuring that the encoder remains general enough for diverse tasks while preserving the EPD model's ability to adapt effectively to specific tasks online.
\begin{wrapfigure}{r}{0.3\textwidth}
    \centering
    \vspace{-1.7cm}
    \includegraphics[width = 0.3\columnwidth]{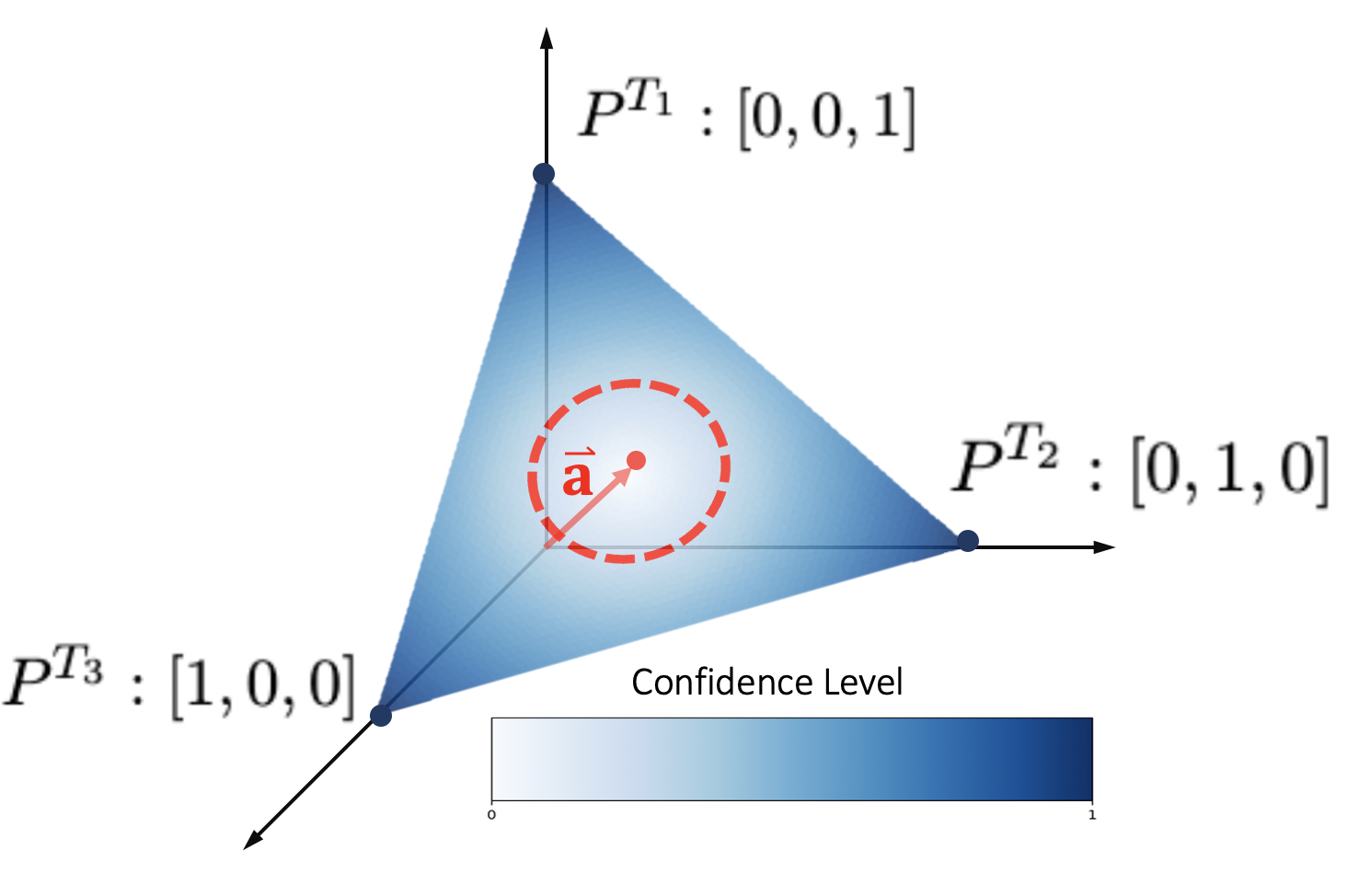}
    \caption{Illustration of the statistical model of task distribution. 
    }
    \vspace{-1.0cm}
    \label{fig: probability simplex}
\end{wrapfigure}
\subsection{Encoder Meta-Update based on Prototype Decoder}
\begin{wrapfigure}{r}{0.45\columnwidth}
\vspace{-4mm}
\centering
\includegraphics[width = 0.35\columnwidth]{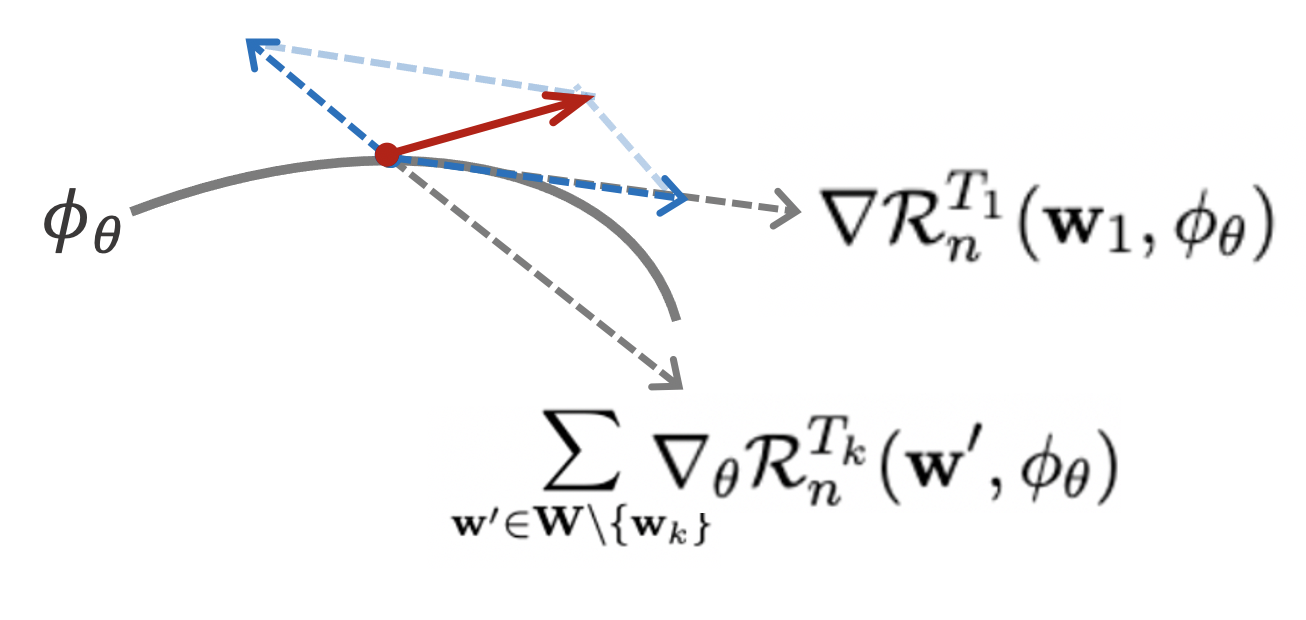}
\vspace{-0.3 cm}
\caption{Illustration of meta update. \textcolor{blue}{Blue} indicates the weighted gradients and \textcolor{red}{red} indicates the update direction}
\label{fig: meta update}
\vspace{-8mm}
\end{wrapfigure}

The prototype decoder is a \textit{local} definition that only represents its corresponding task. The \textit{global} relationships among prototypes are embedded within the encoder in a black-box manner, which determines our ability to understand the underlying task similarities and leverage them for task generalization. To explore the global relationships among the prototypes, we introduce an $N$-dimensional statistical model with the prototype set as a basis in the ``task'' distribution space (see Figure ~\ref{fig: probability simplex}):
\begin{equation}\label{statistical_model}
\mathcal{S}_{\mathbf{W}}(\mathbf{a}) = \{ \sum_{i=1}^N
a_i \mathbf{w}_i \mid \sum_{i=1}^N a_i = 1\ \text{and}\ a_i \geq 0\},
\end{equation}
where $\mathbf{a}=[a_1,\ a_2,\ \ldots, a_N]^T$ is the \textit{coordinates} in the prototype basis, representing the \textit{location} 
of a task distribution in this model. With this model structure~\eqref{statistical_model}, we introduce a prototype-decoder-based meta-update strategy for jointly training the decoders, focusing on exploring the subspace spanned by the prototype basis. The exploration is achieved by adjusting the learning direction through \textbf{negative weighting} of the risk gradients of other tasks' prototypes (see Figure~\ref{fig: meta update}). For task $T_k \in \mathcal{T}$, the one-step meta update is given by:
\begin{equation}\label{regularized_meta_update}
        \theta \leftarrow \theta - \epsilon \Big((1-\beta) \nabla_{\theta} \mathcal{R}^{T_k}_n(\mathbf{w}_k, \phi_{\theta}) - \beta \sum_{\mathbf{w'} \in \mathbf{W}\setminus\{\mathbf{w}_k\}} \nabla_{\theta} \mathcal{R}^{T_k}_n(\mathbf{w'}, \phi_{\theta}) \Big),
\end{equation}
where $\beta \in [0,1)$ is a trade-off parameter, balancing task-specific learning and inter-task interpolation, and $\epsilon$ is the learning rate. In the case of $\beta = 0$, the task-specific prototype remains highly representative of its corresponding task, yet this choice restricts the interpolation on the statistical model \eqref{statistical_model}. Increasing the value of $\beta$ broadens the model's interpolation and coverage during the learning phase but will degrade the representativeness in the sense of a higher risk for the given task. The selection of $\beta$ should align with specific performance metrics: a smaller $\beta$ for concentrated representation to trained tasks and a larger $\beta$ for better extrapolation to new tasks.

The meta update is reminiscent of  \textit{gradient manipulation} in the multi-task learning~\citep{maninis2019attentive, liu2021conflict,liu2021towards}, which aims to balance the learning quality between task-shared and task-specific representations. Note that this balance is explicitly addressed by our EPD structure. Here,  the use of adversarial gradient regularization is specifically designed to explore the vicinity of a given task by introducing tendencies towards other tasks. 


\subsection{\textit{Proto}-MPC}
The EPD model offers an adaptation strategy when used with the MPC to handle uncertainties or disturbances associated with tasks. If privileged information about a task is available online, then \textit{Proto}-MPC can utilize a task-specific residual dynamics model provided by the prototype decoder. Otherwise, in scenarios where task information is not immediately available, we can use prototype decoders to interpolate online data to infer a residual dynamics model.\\

\noindent\textbf{With Privileged Task Information}: Under this condition, MPC can readily choose which model (i.e., prototype decoder) to use. Formally, we describe the task information to be provided by external modules in terms of $\textit{Privileged Information}$ denoted as \texttt{PI} as follows $\mathbf{w}_k = \texttt{PI} ( D^{T_{\text{query}}}_n ),$
where $D^{T_{\text{query}}}_n = \{(x_1, y_1), ..., (x_n, y_n)\}^{T_{\text{query}}}$ is a batch of data from the real-time task $T_{\text{query}}$. This operation essentially outputs the prototype candidate to be used by the MPC.

\noindent\textbf{Without Privileged Task Information}: When task information is not immediately available during operation, the statistical model $\mathcal{S}_{\mathbf{W}}(\mathbf{a})$ (with prototype decoders as a basis) enables a more efficient sampling-based real-time adaptation strategy than recursively solving the empirical risk minimization. Intuitively, this strategy sequentially \textit{locates} the operational task $T_{\text{query}}$ in the (sub)space spanned by the prototype decoders. 

Different from the offline learning stage, we shift our focus from exploration to exploitation at the stage of online adaptation. For exploitation, a challenge comes from the center region in $\mathcal{S}_{\mathbf{W}}(\mathbf{a})$ (see Figure~\ref{fig: probability simplex}) being a \textit{low-confidence} region which is poorly represented in the training data. In particular, the point $\mathbf{a}^* = [\frac{1}{N}\ \frac{1}{N}\ ...\ \frac{1}{N}]$ at the center of $\mathcal{S}_{\mathbf{W}}(\mathbf{a})$ represents the state of highest uncertainty, where each task is equally probable.
To address this challenge, we propose a \textbf{prototype-based coordinates sampling method with an acceptance criterion}, which sequentially updates $\mathbf{a}$ in the high confidence region of $\mathcal{S}_{\mathbf{W}}(\mathbf{a})$.

For the prototype coordinate $\mathbf{a}$, its $k$th element $\mathbf{a}_k$ has a probabilistic interpretation as the probability of the task $T_{\text{query}}$ being $T_k$, i.e., $\mathbf{a}_k = P(T_{\text{query}} = T_k)$. Therefore, the coordinate $\mathbf{a}$ essentially gives the probability distribution of $T_{\text{query}}$ over task set $\mathcal{T}$. In practice, given $D^{T_{\text{query}}}_n$, we can empirically approximate $\mathbf{a}_k$ using Boltzmann distribution:
\begin{equation}\label{eq:empirical a}
    \mathbf{a}_k = P(T_{\text{query}}=T_k) \approx P_{emp}(T_{\text{query}}=T_k) = \mathbf{a}_{emp,k} = \frac{\exp \big( - \gamma \mathcal{R}^{T_{\text{query}}}_n(\mathbf{w}_k, \phi_{\theta}) \big) } {\sum_{\mathbf{w'}\in\mathbf{W}} \exp \big( - \gamma \mathcal{R}_n^{T_{\text{query}}}(\mathbf{w'}, \phi_{\theta})\big)},
\end{equation}
where $\gamma > 0$ is a scaling parameter that controls the weighting to the risk (i.e., a lower value of $\gamma$ tends to ``flatten out'' $P_{emp}$). To keep $\mathbf{a}_{emp}$  away from the highest uncertain point $\mathbf{a}^*$, we define an \textit{acceptance} criterion using Kullback–Leibler divergence with a predefined acceptance threshold $D_0$, i.e., if the following inequality holds $D_{KL} (\mathbf{a}_{emp}\|\mathbf{a}^*) > D_0$,
then $ \mathbf{a}_{emp}$ is considered as bounded away from  $\mathbf{a}^*$ and will be accepted. In the inference stage, $\mathbf{a}_{emp}$ can be recursively computed using a \textit{moving horizon data buffer} to sequentially update the decoder weights online while the acceptance criterion 
ensures that $\mathbf{a}_{emp}$ stays away from the central low-confidence region. We summarize the adaptation scheme of \textit{Proto}-MPC in Algorithm~\ref{alg:proto MPC}. The block diagram of \textit{Proto}-MPC for controlling the quadrotor is illustrated in Fig.~\ref{fig: Framework overview}d.

\section{Experiments}
In this section, we evaluate our method in simulation. We use the RotorPy simulator~\citep{folk2023rotorpy}, a multirotor simulation environment with aerodynamic wrenches, to collect data for training the EPD model and test the \textit{Proto}-MPC.\\

\noindent\textbf{Experimental Setup:} The learning task set is designed for constant side wind in the $x$-direction at speeds of 2, 4, and 6 m/s. In this scenario, the lumped forces dominate the residual dynamics $f_{\Delta}$. Therefore, only the lumped forces are considered in the learned residual dynamics of this experimental setup. See the Appendix for details on the MPC implementation, data collection, and training results of the EPD model.\\

\begin{wrapfigure}{r}{0.4\textwidth}
    \centering
    \vspace{-0.8cm}
    \includegraphics[width = 0.4\columnwidth]{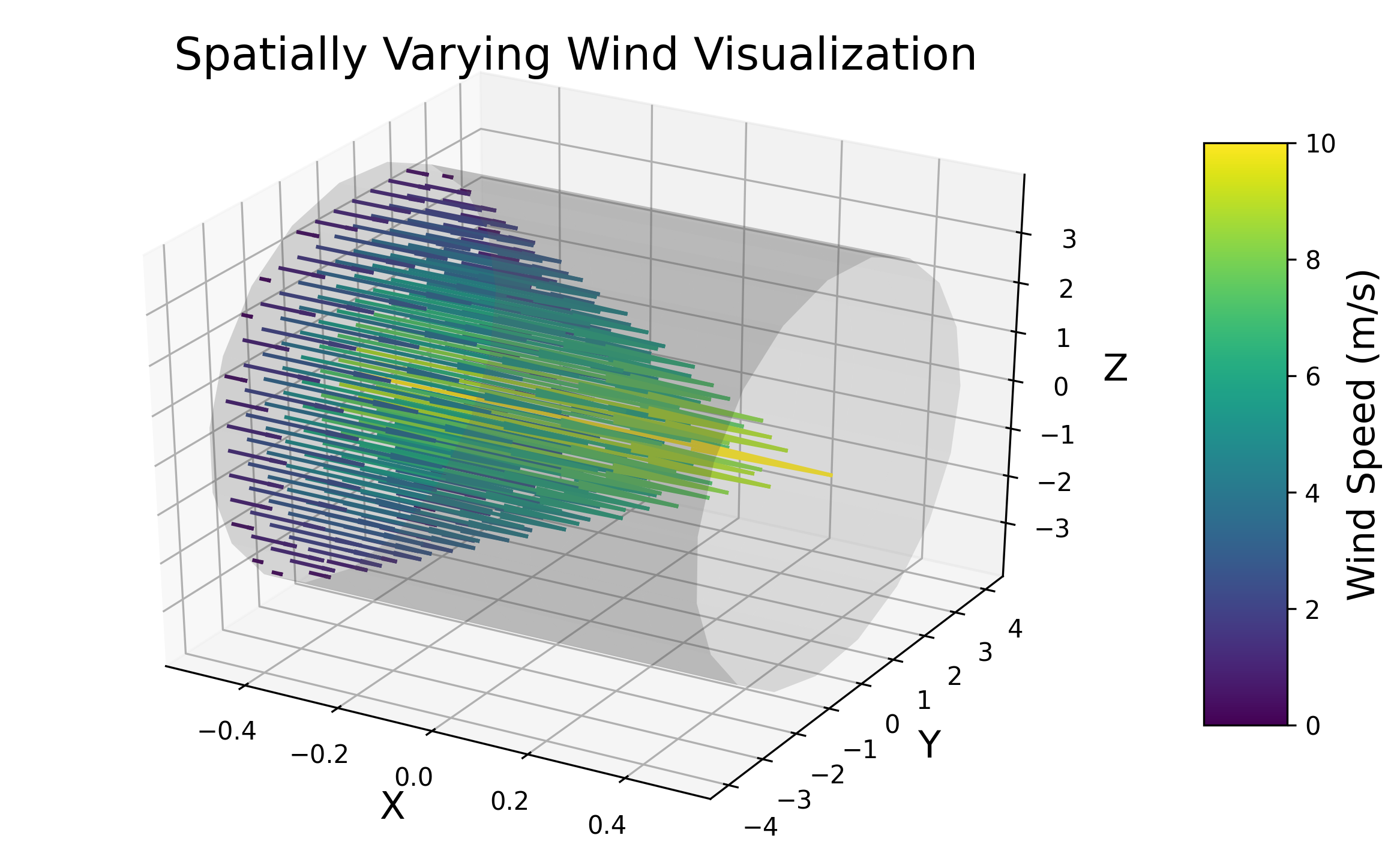}
    \vspace{-6mm} 
    \caption{Spatially varying wind distribution.
    }
    \label{fig: wind}
     \vspace{-5mm} 
\end{wrapfigure}

\noindent\textbf{Experimental Results}: To evaluate our method, we compare it with 1) nonlinear MPC with nominal model $f_{\text{nom}}$, 2) KNODE-MPC-Online ~\citep{jiahao2023online}, and 3) MPC with task-specific DNN residual model ($f^{T_k}_{\theta}$ is a DNN trained using the $T_k$-specific dataset). In other words, for each task, a DNN residual model is trained and used for deployment on the given task. On the contrary, for all the testing trials we conduct in this subsection, the prototype-based meta-model is \textbf{kept fixed} so that the adaptation is only based on prototype decoders corresponding to constant side wind with speeds of 2, 4, and 6 m/s. We evaluate our method under static and dynamic wind scenarios.
For the former, we command the quadrotors to track the training trajectory under constant side wind of various speeds. For the latter, we command the quadrotors to track different testing trajectories under spatially dependent winds (0--10 m/s along the $x$-direction; see the illustration in Figure~\ref{fig: wind}).\\

\noindent \textit{Constant Side Wind}: Table~\ref{tb: rmse ref trajectory} presents a comparison of tracking RMSE for nominal MPC, task-specific DNN-MPC, KNODE-MPC-Online, and \textit{Proto}-MPC under constant side wind conditions with speeds ranging from 0 to 10 m/s. We followed the implementation of KNODE-MPC-Online as described in~\citep{jiahao2023online} for handling sudden mass changes online but adapted it for our side wind setup. Empirically, we found that the original implementation suffers from instability issues with the online learned model in our experimental setup. To address this, we applied spectral normalization to control the Lipschitz constant of the online-learned KNODE model, thereby improving its closed-loop stability. 


The result shows a substantial reduction in RMSE for all task-specific DNN-MPC, KNODE-MPC-Online and \textit{Proto}-MPC compared to the baseline MPC. Note that the task-specific DNN-MPC is expected to exhibit superior tracking performance, as the DNN is specifically trained for each task's wind condition. Both KNODE-MPC-Online and \textit{Proto}-MPC consistently halve the RMSE relative to the nominal MPC across all test wind speeds. However, \textit{Proto}-MPC requires less online computation compared to KNODE-MPC-Online, as it only updates the decoders instead of training the whole model online. This comparison demonstrates not only a significant improvement over the baseline MPC but also demonstrates \textit{Proto}-MPC's robust generalization capabilities on tasks unseen during training, with significantly lower computational demands.\\

\setlength{\tabcolsep}{5pt} 
\renewcommand{\arraystretch}{1} 
  \captionsetup{
	skip=5pt, position = bottom}
\begin{table}[h]
	\centering
	\small
	
	\vspace{-0.2cm}
	\captionsetup{font=small}
	\caption{Tracking RMSE on the training trajectory (shown in Figure~\ref{fig: reference trajectory}) under constant side winds of different speeds. The bold font for 2, 4, and 6 m/s cases indicate the wind speeds for the training tasks.}
	\begin{tabular}{ccccccccc}
		\toprule[1pt]
		RMSE[m]  & axis  & 0 m/s & \textbf{2 m/s} & \textbf{4 m/s} & \textbf{6 m/s} & 8 m/s & 10 m/s
		\\
		\midrule
        \multirowcell{3}{nominal-MPC} & $x$ & 0.10 & 0.15 & 0.24 & 0.36 & 0.48 & 0.63
  \\
   & $y$ & 0.07 & 0.07 & 0.08 & 0.08 & 0.09 & 0.10
  \\
   & $z$ & 0.03 & 0.03 & 0.05 & 0.08 & 0.12 & 0.16
  \\
  \midrule
  \multirowcell{3}{Task-DNN-MPC} & $x$ & - & 0.08 & 0.09 & 0.11 & 0.12 & 0.15
  \\
  & $y$ & - & 0.06 & 0.05 & 0.06 & 0.05 & 0.05
  \\
   & $z$ & - & 0.03 & 0.03 & 0.03 & 0.04 & 0.04
  \\
  \midrule
  \multirowcell{3}{KNODE-MPC-Online \\  (with Spectral Normalization)} & $x$ & 0.08 & 0.09 & 0.11 & 0.18 & 0.26 & 0.31
  \\
  & $y$ & 0.05 & 0.07 & 0.07 & 0.13 & 0.16 & 0.11
  \\
   & $z$ & 0.06 & 0.05 & 0.08 & 0.17 & 0.18 & 0.22
  \\
  \midrule
  \multirowcell{3}{\textit{Proto}-MPC \\ (with \texttt{PI})} & $x$ & 0.09 & 0.07 & 0.09 & 0.11 & 0.17 & 0.30
  \\
  & $y$ & 0.04 & 0.04 & 0.05 & 0.05 & 0.05 & 0.05
  \\
   & $z$ & 0.03 & 0.02 & 0.03 & 0.03 &0.03 & 0.04
  \\
  \midrule
  \multirowcell{3}{\textit{Proto}-MPC \\ (without \texttt{PI})} & $x$ & 0.10 & 0.07 & 0.13 & 0.17 & 0.24 & 0.32
  \\
  & $y$ & 0.04 & 0.04 & 0.04 & 0.05 & 0.05 & 0.05
  \\
   & $z$ & 0.03 & 0.02 & 0.02 & 0.03 &0.03 & 0.04
  \\
       	\bottomrule[1pt]
	\end{tabular}\label{tb: rmse ref trajectory}
\end{table}
\normalsize

\begin{figure}[h]
    \floatconts
    {fig: testing results} 
    {\caption{Tracking performance subject to spatially varying winds on different trajectories. The first row (\ref{fig:path 1 mpc}, \ref{fig:path 2 mpc}, \ref{fig:path 3 mpc}) shows the tracking performance of MPC with the nominal model, the second row (\ref{fig:path 1 knode}, \ref{fig:path 2 knode}, \ref{fig:path 3 knode}) shows the tracking performance of KNODE-MPC-Online (with spectral normalization) and the third row (\ref{fig:path 1 proto}, \ref{fig:path 2 proto}, \ref{fig:path 3 proto}) shows the tracking performance of \textit{Proto}-MPC. The colorbar highlights the deviation from the reference trajectory}}
    {
    \subfigure[][c]
    {
    \label{fig:path 1 mpc} 
    \includegraphics[width = 0.28\columnwidth]{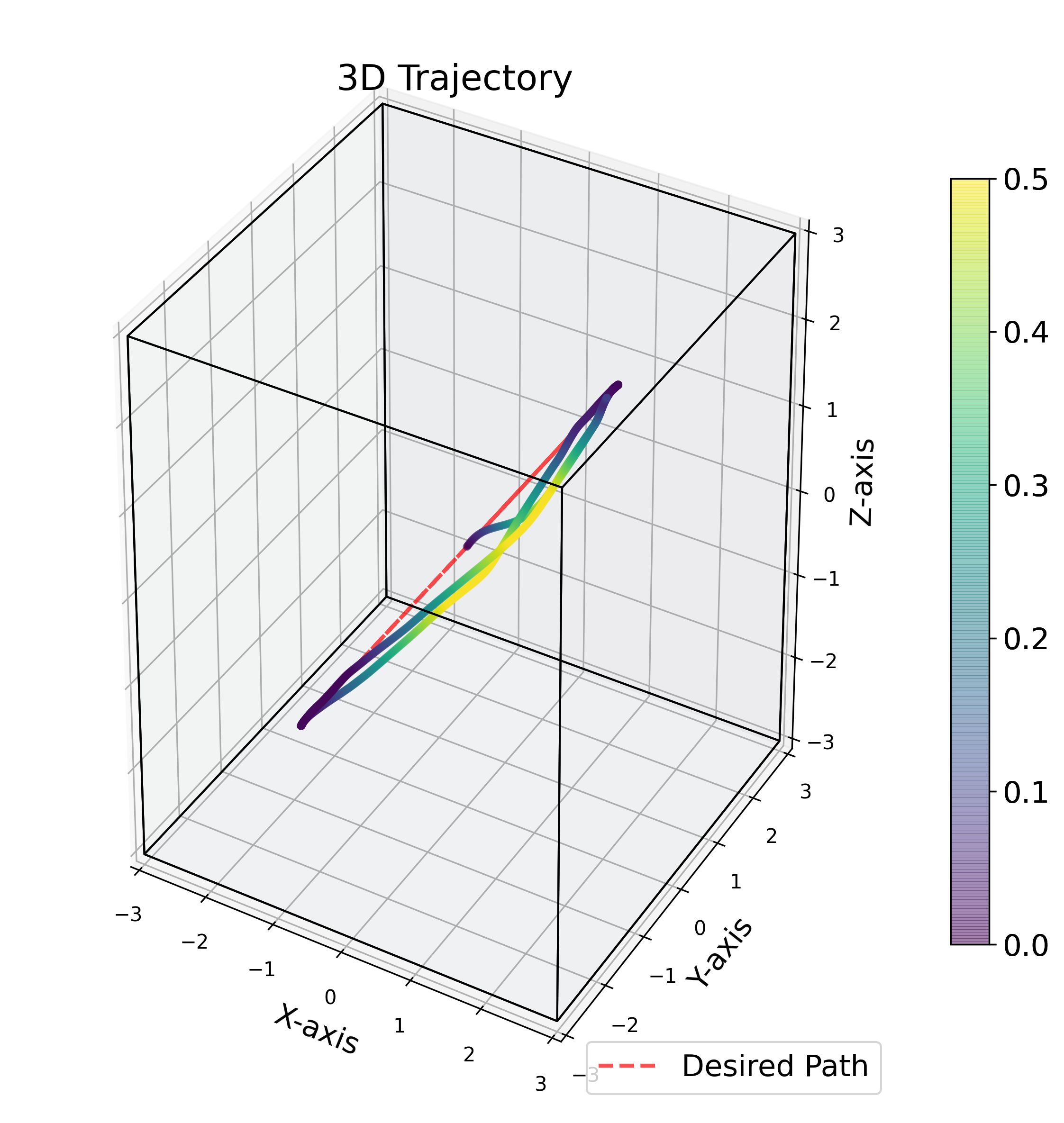} }
    \subfigure[][c]
    {
    \label{fig:path 2 mpc} 
    \includegraphics[width = 0.28\columnwidth]{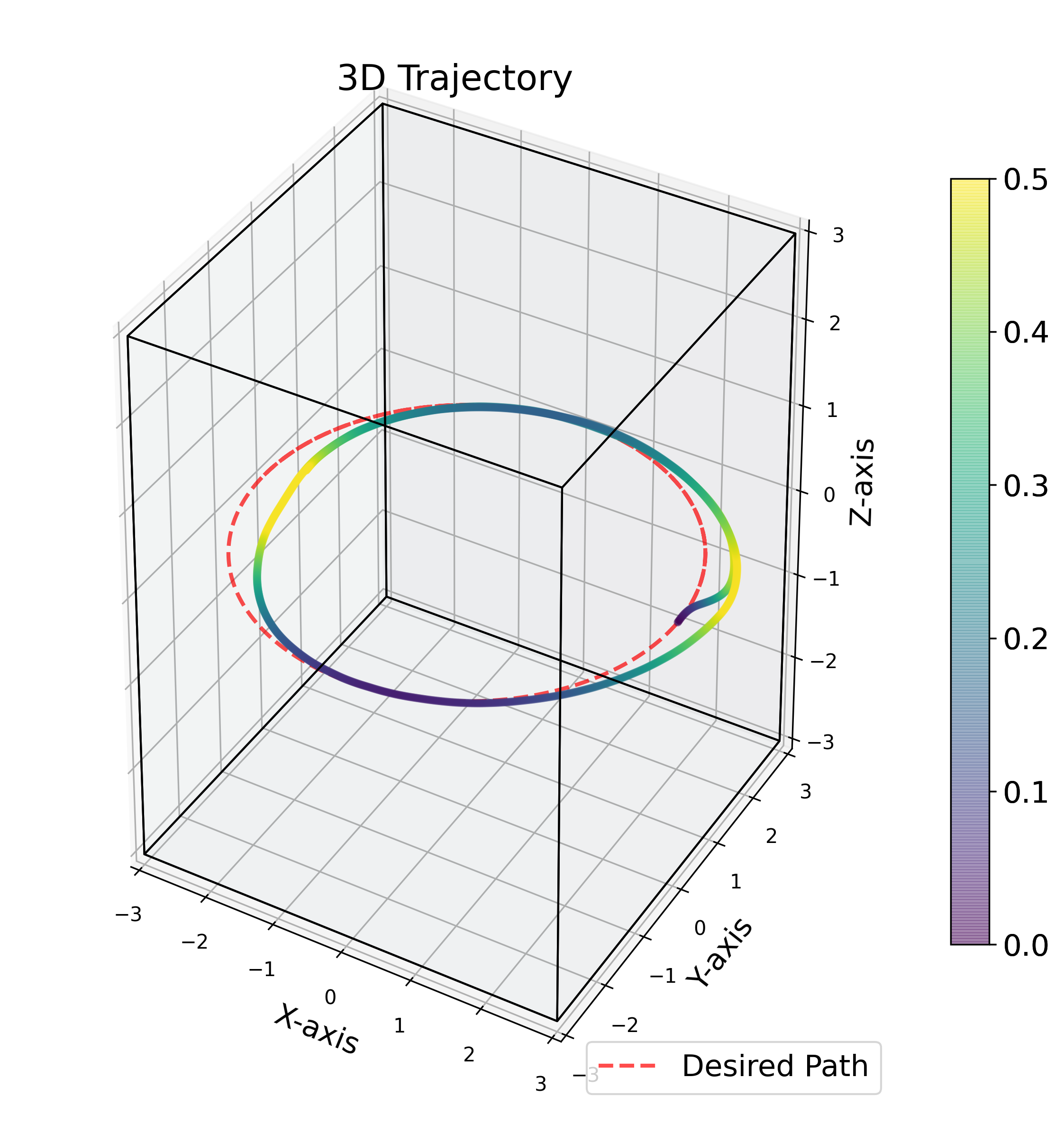} }
    \subfigure[][c]
    { 
    \label{fig:path 3 mpc}
    \includegraphics[width = 0.28\columnwidth]{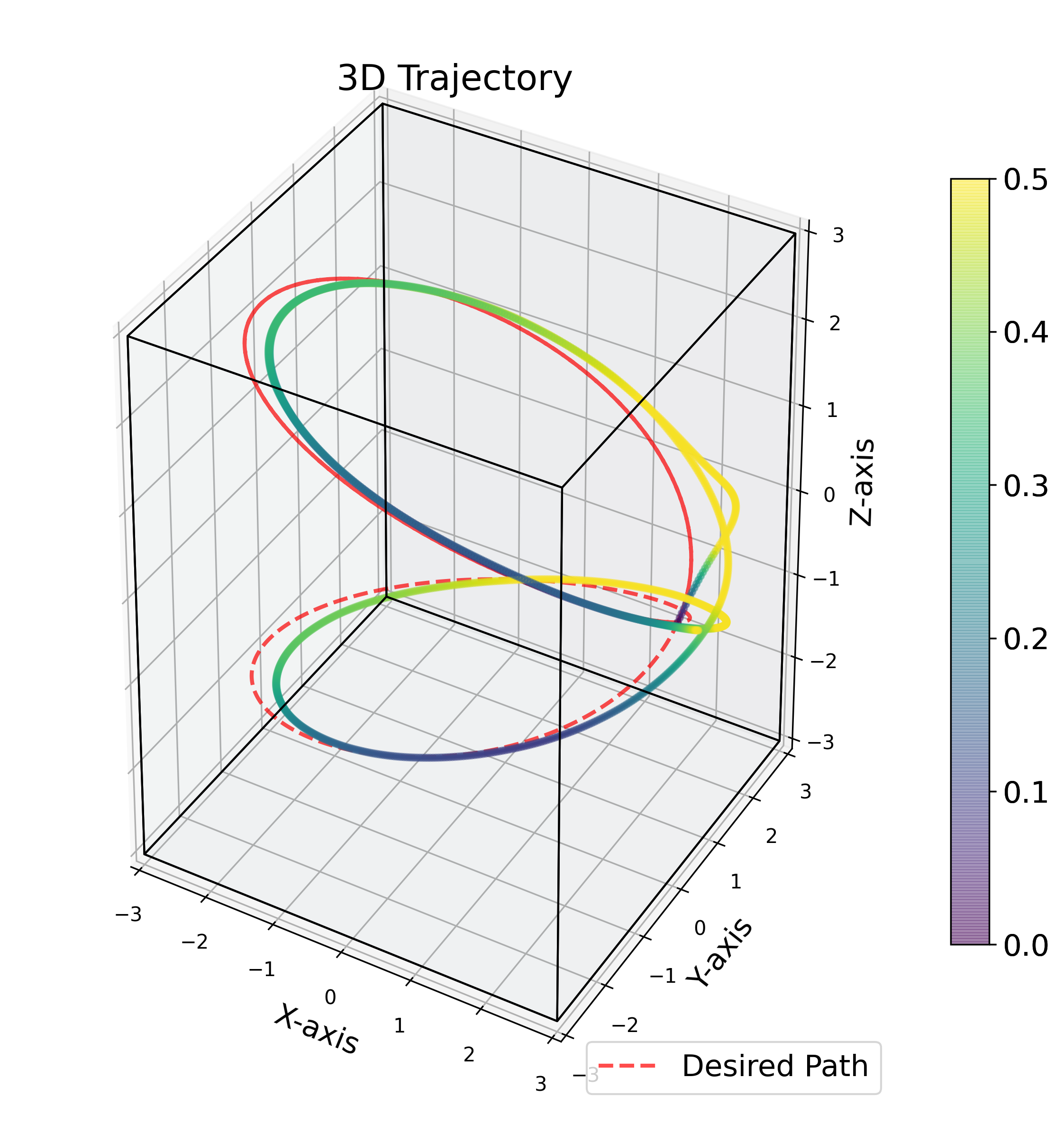}
    }
    \subfigure[][c]
    {
    \label{fig:path 1 knode} 
    \includegraphics[width = 0.28\columnwidth]{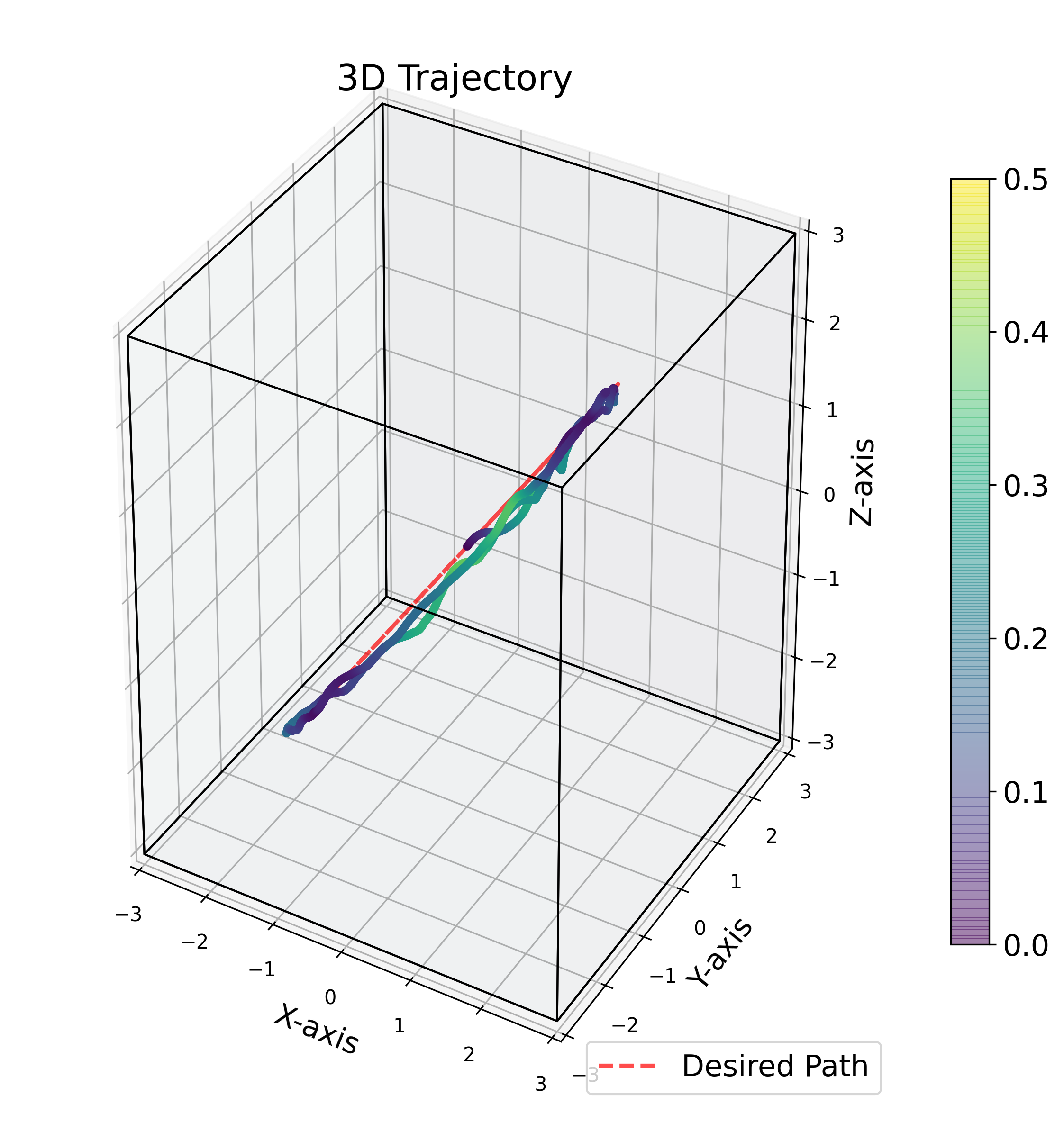} }
    \subfigure[][c]
    {
    \label{fig:path 2 knode} 
    \includegraphics[width = 0.28\columnwidth]{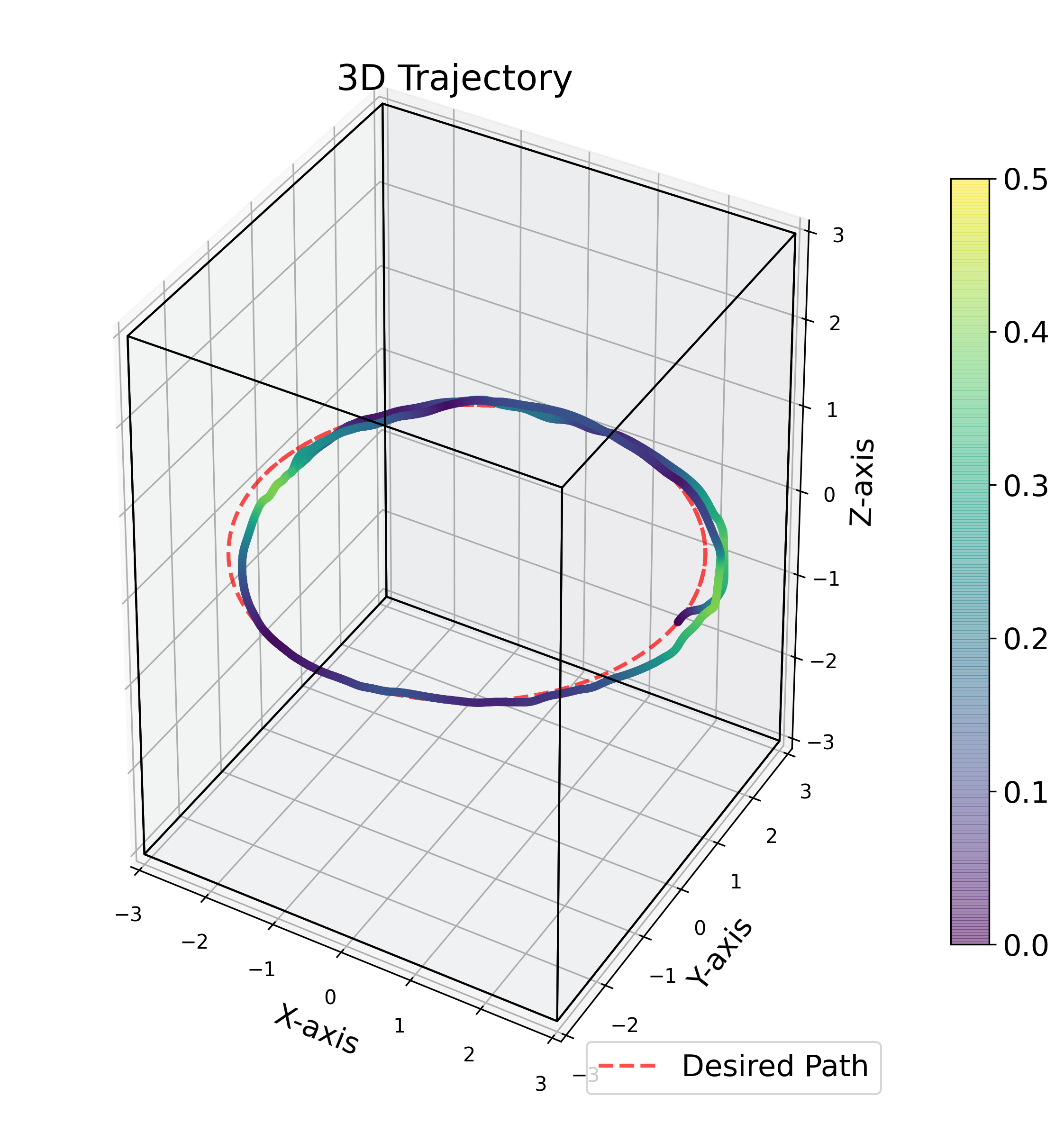} }
    \subfigure[][c]
    { 
    \label{fig:path 3 knode}
    \includegraphics[width = 0.28\columnwidth]{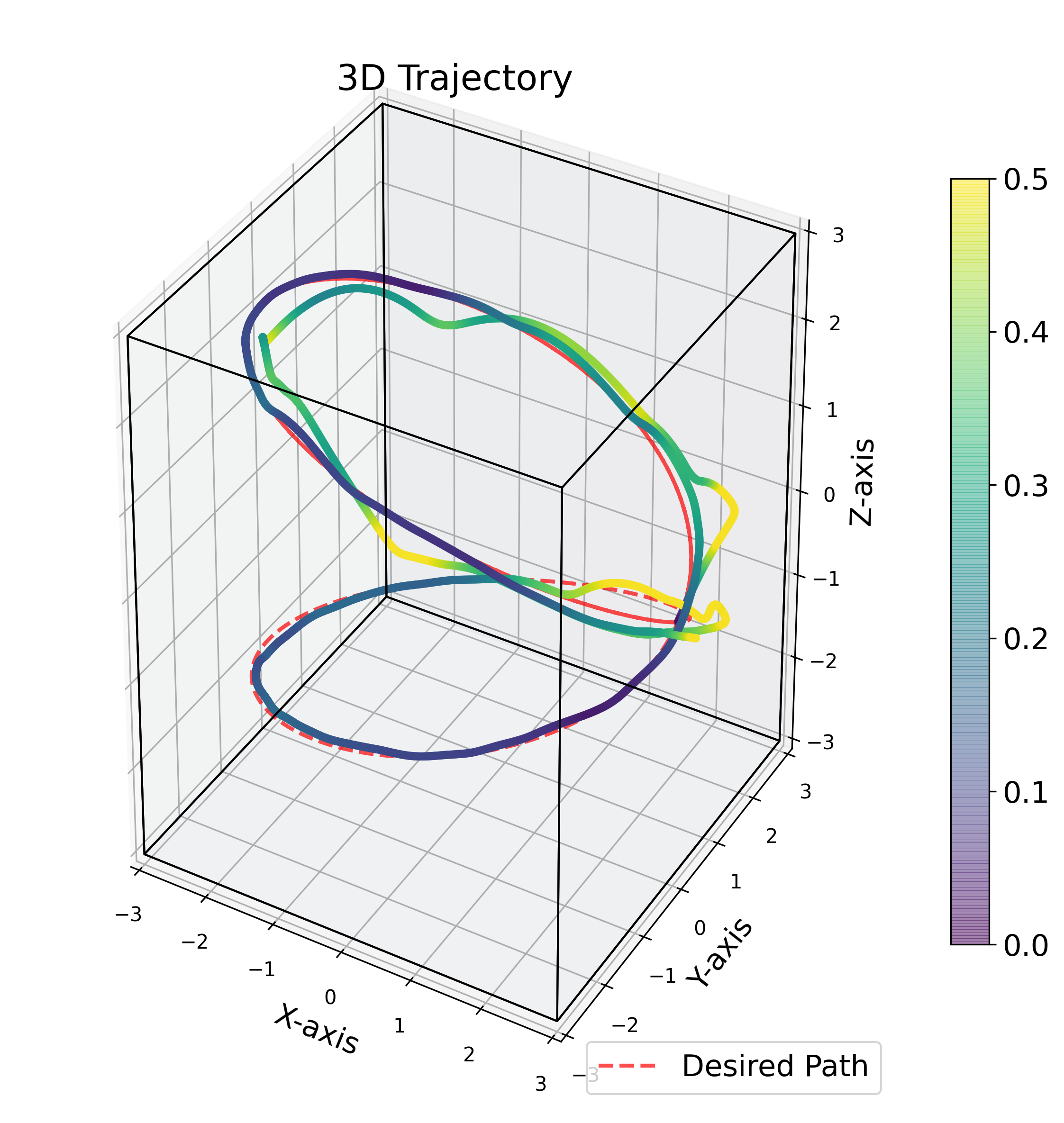}
    }
    \subfigure[][c]
    {
    \label{fig:path 1 proto} 
    \includegraphics[width = 0.28\columnwidth]{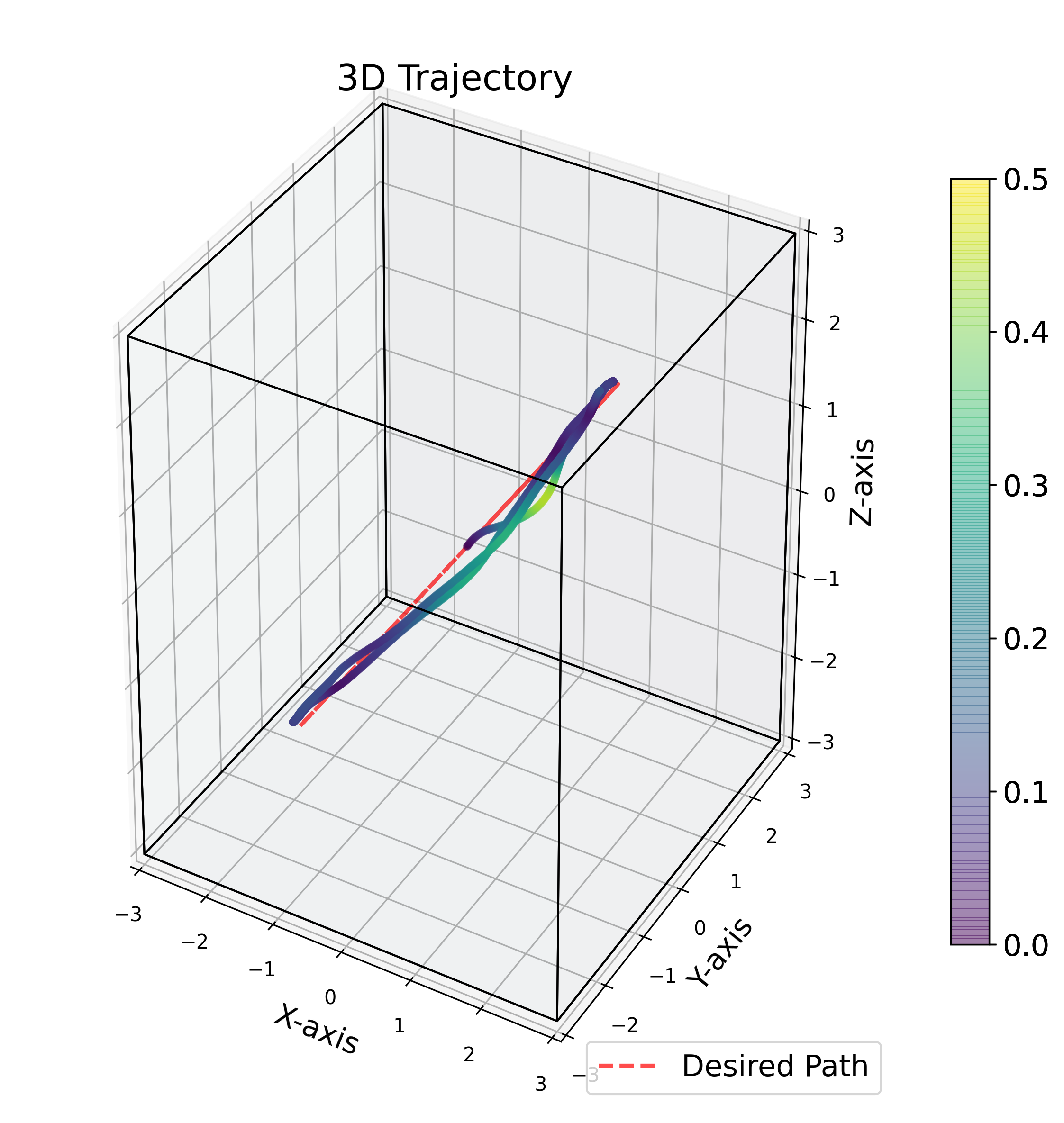} }
    \subfigure[][c]
    {
    \label{fig:path 2 proto} 
    \includegraphics[width = 0.28\columnwidth]{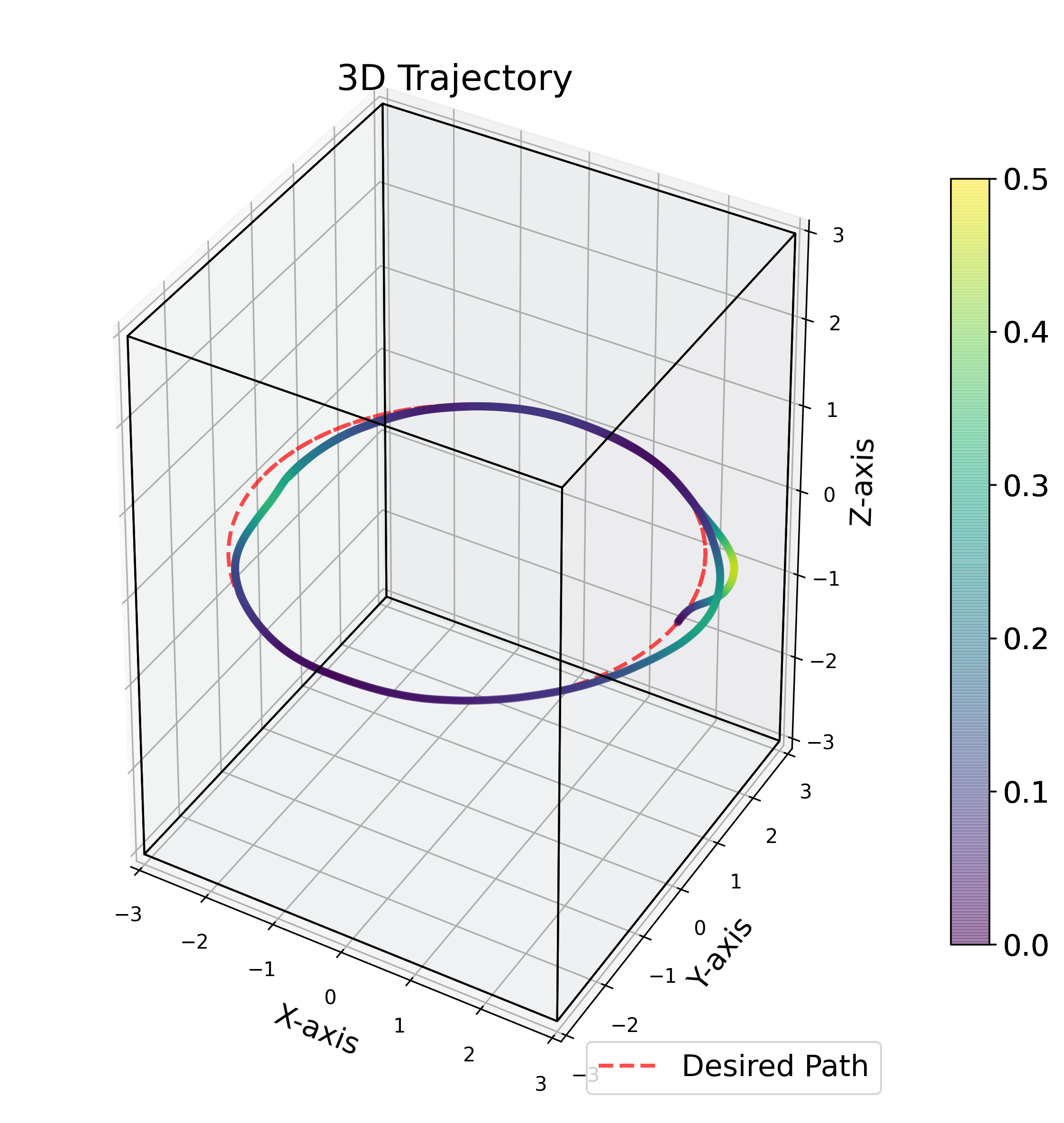} }
    \subfigure[][c]
    { 
    \label{fig:path 3 proto}
    \includegraphics[width = 0.28\columnwidth]{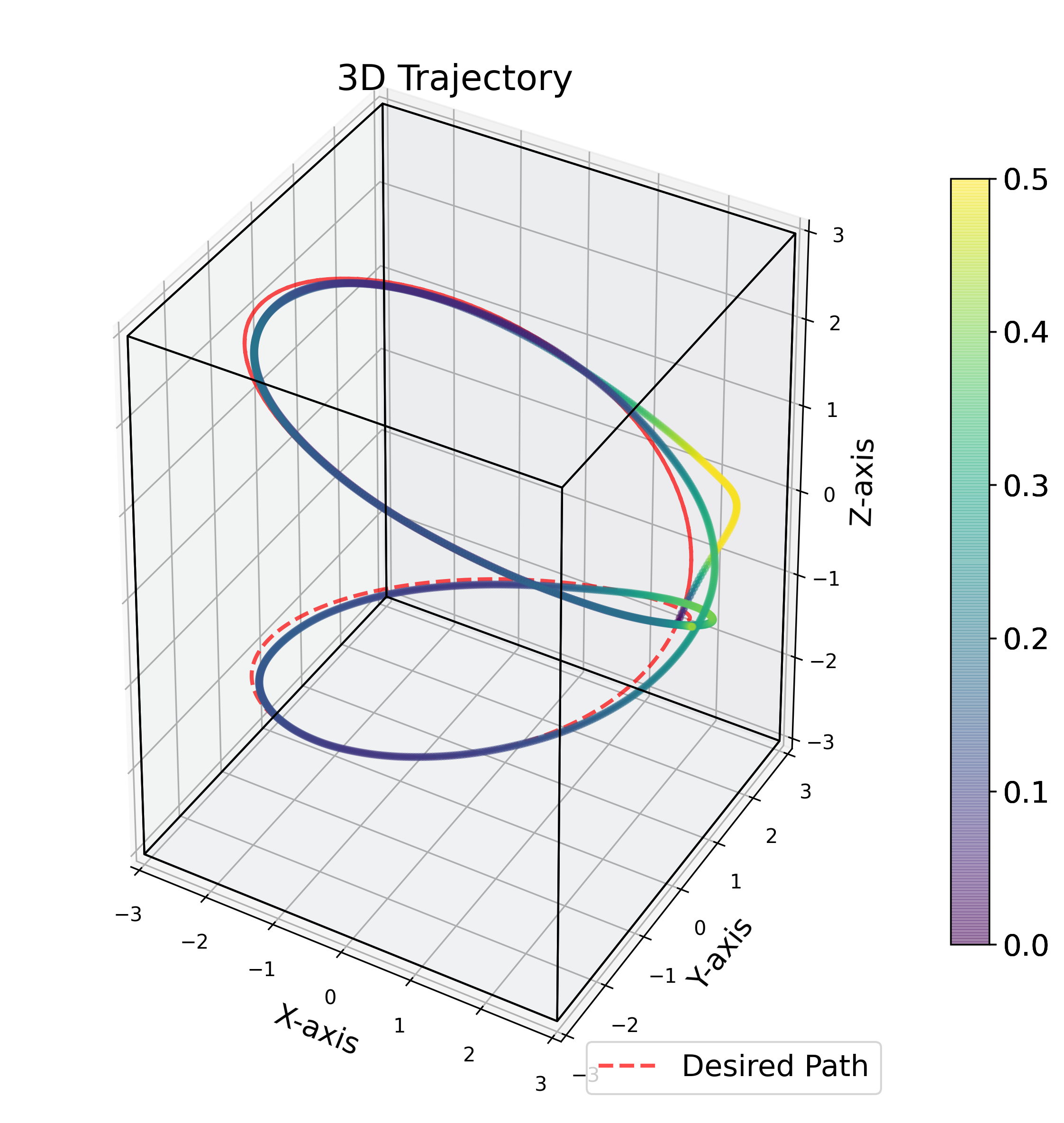}
    }
    }
    \vspace{-0.3 cm}
    \label{fig: 3d path testing}
\end{figure}

\noindent \textit{Spatially Varying Wind}: Under this condition, to test \textit{Proto}-MPC's task-adaptation capacity, the quadrotor is subject to a varying-speed wind in the $x$-direction from 0 to 10 m/s. We compare it with nominal-MPC and KNODE-MPC-Online (with spectral normalization) on various trajectories. Figure~\ref{fig: 3d path testing} shows the tracking performance with the colorbar highlighting the deviation from the reference trajectory. Table~\ref{tb: rmse testing trajectories} shows the RMSE of the three methods on the testing trajectories (the associated box plot is attached in the Appendix, see Fig.~\ref{fig: 3d path error distribution}.). Compared with nominal MPC and KNODE-MPC-online, the \textit{Proto}-MPC achieves the best trajectory tracking under drastically changing wind conditions with significantly less online computation.
\setlength{\tabcolsep}{5pt} 
\renewcommand{\arraystretch}{1} 
  \captionsetup{
	skip=5pt, position = bottom}
\begin{table}[h]
	\centering
	\small
	
	\vspace{-0.2cm}
	\captionsetup{font=small}
	\caption{Tracking RMSE on the testing trajectories (shown in Figure~\ref{fig: 3d path testing}) under spatially varying wind.}
	\begin{tabular}{cccccc}
		\toprule[1pt]
		RMSE[m]  & axis  & trajectory 1 & trajectory 2 & trajectory 3 
		\\
		\midrule
        \multirowcell{3}{nominal-MPC} & $x$ & 0.25 & 0.31 & 0.35 
  \\
   & $y$ & 0.05 & 0.06 & 0.11 
  \\
   & $z$ & 0.06 & 0.06 & 0.09
  \\
  \midrule
  \multirowcell{3}{KNODE-MPC-Online \\ (with Spectral Normalization)} & $x$ & 0.15 & 0.17 & 0.22 
  \\
  & $y$ & 0.06 & 0.05 & 0.09
  \\
   & $z$ & 0.09 & 0.08 & 0.06
  \\
  \midrule
  \multirowcell{3}{\textit{Proto}-MPC \\ (without \texttt{PI})} & $x$ & 0.12 & 0.15 & 0.18 
  \\
  & $y$ & 0.03 & 0.03 & 0.12
  \\
   & $z$ & 0.02 & 0.02 & 0.08
  \\
       	\bottomrule[1pt]
	\end{tabular}
 \label{tb: rmse testing trajectories}
\end{table}
\normalsize

\section{Conclusion}
This paper proposes a novel EPD model designed to capture shared and distinctive features across various training tasks. The EPD model consists of a universal task-agnostic DNN encoder and a set of task-specific linear prototype decoders to balance task-shared and task-specific representations. In the online setting, the encoder processes incoming data into features. Simultaneously, the linear prototype decoders are used as a ``basis'' to interpolate encoded features, which allows fast computation
of a new decoder aligned with the current task’s characteristics. 
We then use the EPD model to capture residual dynamics in our \textit{Proto}-MPC, which can quickly adapt the model to cope with uncertainties from dynamically evolving task scenarios. We evaluate \textit{Proto}-MPC's performance in controlling a quadrotor to track agile trajectories under various static and dynamic side wind conditions, which demonstrates its robust performance compared to nominal MPC and its generalization capacity compared to MPC augmented with task-specific DNN residual models.  Future directions include deploying this framework in real-world experiments and investigate how the geometric properties of prototype decoders help to better understand the underlying relationships between tasks on the manifold.

\acks{This work is supported by NASA under the Cooperative Agreement 80NSSC20M0229 and University Leadership Initiative grant 80NSSC22M0070, NSF-AoF Robust Intelligence award \#2133656, NSF SLES \#2331878, and DoD HQ00342110002.}

\bibliography{ref}

\begin{thebibliography}{26}
\providecommand{\natexlab}[1]{#1}
\providecommand{\url}[1]{\texttt{#1}}
\expandafter\ifx\csname urlstyle\endcsname\relax
  \providecommand{\doi}[1]{doi: #1}\else
  \providecommand{\doi}{doi: \begingroup \urlstyle{rm}\Url}\fi

\bibitem[Arimoto(1972)]{arimoto1972algorithm}
Suguru Arimoto.
\newblock An algorithm for computing the capacity of arbitrary discrete memoryless channels.
\newblock \emph{IEEE Transactions on Information Theory}, 18\penalty0 (1):\penalty0 14--20, 1972.

\bibitem[Chee et~al.(2022)Chee, Jiahao, and Hsieh]{chee2022knode}
Kong~Yao Chee, Tom~Z Jiahao, and M~Ani Hsieh.
\newblock {KNODE-MPC}: A knowledge-based data-driven predictive control framework for aerial robots.
\newblock \emph{IEEE Robotics and Automation Letters}, 7\penalty0 (2):\penalty0 2819--2826, 2022.

\bibitem[Chen et~al.(2018)Chen, Rubanova, Bettencourt, and Duvenaud]{chen2018neural}
Ricky~TQ Chen, Yulia Rubanova, Jesse Bettencourt, and David~K Duvenaud.
\newblock Neural ordinary differential equations.
\newblock \emph{Advances in neural information processing systems}, 31, 2018.

\bibitem[Cover(1999)]{cover1999elements}
Thomas~M Cover.
\newblock \emph{Elements of information theory}.
\newblock John Wiley \& Sons, 1999.

\bibitem[Diehl et~al.(2006)Diehl, Bock, Diedam, and Wieber]{diehl2006fast}
Moritz Diehl, Hans~Georg Bock, Holger Diedam, and P-B Wieber.
\newblock Fast direct multiple shooting algorithms for optimal robot control.
\newblock \emph{Fast motions in biomechanics and robotics: optimization and feedback control}, pages 65--93, 2006.

\bibitem[Folk et~al.(2023)Folk, Paulos, and Kumar]{folk2023rotorpy}
Spencer Folk, James Paulos, and Vijay Kumar.
\newblock {RotorPy}: A python-based multirotor simulator with aerodynamics for education and research.
\newblock \emph{arXiv preprint arXiv:2306.04485}, 2023.

\bibitem[Hwangbo et~al.(2017)Hwangbo, Sa, Siegwart, and Hutter]{hwangbo2017control}
Jemin Hwangbo, Inkyu Sa, Roland Siegwart, and Marco Hutter.
\newblock Control of a quadrotor with reinforcement learning.
\newblock \emph{IEEE Robotics and Automation Letters}, 2\penalty0 (4):\penalty0 2096--2103, 2017.

\bibitem[Jiahao et~al.(2023)Jiahao, Chee, and Hsieh]{jiahao2023online}
Tom~Z Jiahao, Kong~Yao Chee, and M~Ani Hsieh.
\newblock Online dynamics learning for predictive control with an application to aerial robots.
\newblock In \emph{Proceedings of the Conference on Robot Learning}, pages 2251--2261. PMLR, 2023.

\bibitem[Joshi et~al.(2021)Joshi, Virdi, and Chowdhary]{joshi2021asynchronous}
Girish Joshi, Jasvir Virdi, and Girish Chowdhary.
\newblock Asynchronous deep model reference adaptive control.
\newblock In \emph{Proceedings of the Conference on Robot Learning}, pages 984--1000. PMLR, 2021.

\bibitem[Kabzan et~al.(2019)Kabzan, Hewing, Liniger, and Zeilinger]{kabzan2019learning}
Juraj Kabzan, Lukas Hewing, Alexander Liniger, and Melanie~N Zeilinger.
\newblock Learning-based model predictive control for autonomous racing.
\newblock \emph{IEEE Robotics and Automation Letters}, 4\penalty0 (4):\penalty0 3363--3370, 2019.

\bibitem[Lambert et~al.(2019)Lambert, Drew, Yaconelli, Levine, Calandra, and Pister]{lambert2019low}
Nathan~O Lambert, Daniel~S Drew, Joseph Yaconelli, Sergey Levine, Roberto Calandra, and Kristofer~SJ Pister.
\newblock Low-level control of a quadrotor with deep model-based reinforcement learning.
\newblock \emph{IEEE Robotics and Automation Letters}, 4\penalty0 (4):\penalty0 4224--4230, 2019.

\bibitem[Liu et~al.(2021{\natexlab{a}})Liu, Liu, Jin, Stone, and Liu]{liu2021conflict}
Bo~Liu, Xingchao Liu, Xiaojie Jin, Peter Stone, and Qiang Liu.
\newblock Conflict-averse gradient descent for multi-task learning.
\newblock \emph{Advances in Neural Information Processing Systems}, 34:\penalty0 18878--18890, 2021{\natexlab{a}}.

\bibitem[Liu et~al.(2021{\natexlab{b}})Liu, Li, Kuang, Xue, Chen, Yang, Liao, and Zhang]{liu2021towards}
Liyang Liu, Yi~Li, Zhanghui Kuang, J~Xue, Yimin Chen, Wenming Yang, Qingmin Liao, and Wayne Zhang.
\newblock Towards impartial multi-task learning.
\newblock In \emph{Proceedings of the International Conference on Learning Representations}, 2021{\natexlab{b}}.

\bibitem[Maninis et~al.(2019)Maninis, Radosavovic, and Kokkinos]{maninis2019attentive}
Kevis-Kokitsi Maninis, Ilija Radosavovic, and Iasonas Kokkinos.
\newblock Attentive single-tasking of multiple tasks.
\newblock In \emph{Proceedings of the IEEE/CVF conference on computer vision and pattern recognition}, pages 1851--1860, 2019.

\bibitem[Mellinger and Kumar(2011)]{mellinger2011minimum}
Daniel Mellinger and Vijay Kumar.
\newblock Minimum snap trajectory generation and control for quadrotors.
\newblock In \emph{Proceedings of the 2011 IEEE International Conference on Robotics and Automation}, pages 2520--2525. IEEE, 2011.

\bibitem[O’Connell et~al.(2022)O’Connell, Shi, Shi, Azizzadenesheli, Anandkumar, Yue, and Chung]{o2022neural}
Michael O’Connell, Guanya Shi, Xichen Shi, Kamyar Azizzadenesheli, Anima Anandkumar, Yisong Yue, and Soon-Jo Chung.
\newblock Neural-fly enables rapid learning for agile flight in strong winds.
\newblock \emph{Science Robotics}, 7\penalty0 (66):\penalty0 eabm6597, 2022.

\bibitem[Richards et~al.(2021)Richards, Azizan, Slotine, and Pavone]{richards2021adaptive}
SM~Richards, N~Azizan, J-JE Slotine, and M~Pavone.
\newblock Adaptive-control-oriented meta-learning for nonlinear systems.
\newblock In \emph{Robotics science and systems}, 2021.

\bibitem[Saviolo et~al.(2022)Saviolo, Li, and Loianno]{saviolo2022physics}
Alessandro Saviolo, Guanrui Li, and Giuseppe Loianno.
\newblock Physics-inspired temporal learning of quadrotor dynamics for accurate model predictive trajectory tracking.
\newblock \emph{IEEE Robotics and Automation Letters}, 7\penalty0 (4):\penalty0 10256--10263, 2022.

\bibitem[Saviolo et~al.(2023)Saviolo, Frey, Rathod, Diehl, and Loianno]{saviolo2023active}
Alessandro Saviolo, Jonathan Frey, Abhishek Rathod, Moritz Diehl, and Giuseppe Loianno.
\newblock Active learning of discrete-time dynamics for uncertainty-aware model predictive control.
\newblock \emph{IEEE Transactions on Robotics}, 2023.

\bibitem[Snell et~al.(2017)Snell, Swersky, and Zemel]{snell2017prototypical}
Jake Snell, Kevin Swersky, and Richard Zemel.
\newblock Prototypical networks for few-shot learning.
\newblock \emph{Advances in neural information processing systems}, 30, 2017.

\bibitem[Tishby et~al.(2000)Tishby, Pereira, and Bialek]{tishby2000information}
Naftali Tishby, Fernando~C Pereira, and William Bialek.
\newblock The information bottleneck method.
\newblock \emph{arXiv preprint physics/0004057}, 2000.

\bibitem[Torrente et~al.(2021)Torrente, Kaufmann, F{\"o}hn, and Scaramuzza]{torrente2021data}
Guillem Torrente, Elia Kaufmann, Philipp F{\"o}hn, and Davide Scaramuzza.
\newblock Data-driven {MPC} for quadrotors.
\newblock \emph{IEEE Robotics and Automation Letters}, 6\penalty0 (2):\penalty0 3769--3776, 2021.

\bibitem[Verschueren et~al.(2018)Verschueren, Frison, Kouzoupis, van Duijkeren, Zanelli, Quirynen, and Diehl]{Verschueren2018}
Robin Verschueren, Gianluca Frison, Dimitris Kouzoupis, Niels van Duijkeren, Andrea Zanelli, Rien Quirynen, and Moritz Diehl.
\newblock Towards a modular software package for embedded optimization.
\newblock \emph{IFAC-PapersOnLine}, 51\penalty0 (20):\penalty0 374--380, 2018.

\bibitem[Wang et~al.(2024)Wang, Ma, Lai, and Zhao]{wang2024neural}
Bingheng Wang, Zhengtian Ma, Shupeng Lai, and Lin Zhao.
\newblock Neural moving horizon estimation for robust flight control.
\newblock \emph{IEEE Transactions on Robotics}, 40:\penalty0 639--659, 2024.

\bibitem[Williams et~al.(2017)Williams, Wagener, Goldfain, Drews, Rehg, Boots, and Theodorou]{williams2017information}
Grady Williams, Nolan Wagener, Brian Goldfain, Paul Drews, James~M Rehg, Byron Boots, and Evangelos~A Theodorou.
\newblock Information theoretic mpc for model-based reinforcement learning.
\newblock In \emph{Proceedings of the 2017 IEEE International Conference on Robotics and Automation (ICRA)}, pages 1714--1721. IEEE, 2017.

\bibitem[Wu et~al.(2022)Wu, Cheng, Ackerman, Gahlawat, Lakshmanan, Zhao, and Hovakimyan]{wu20221}
Zhuohuan Wu, Sheng Cheng, Kasey~A Ackerman, Aditya Gahlawat, Arun Lakshmanan, Pan Zhao, and Naira Hovakimyan.
\newblock $\mathcal{L}_1$ adaptive augmentation for geometric tracking control of quadrotors.
\newblock In \emph{Proceedings of the 2022 International Conference on Robotics and Automation (ICRA)}, pages 1329--1336. IEEE, 2022.

\end{thebibliography}

\newpage
\section{Appendix}
\subsection{Related Work}
We give a brief overview of learning-based control methods with a focus on their applications to quadrotors. ~\citep{hwangbo2017control} uses model-free reinforcement learning to train an end-to-end neural network-based control policy to stabilize a quadrotor under challenging
initial poses (i.e., upside-down). In contrast to end-to-end methods, ~\citep{lambert2019low} learns a deep neural network (DNN) dynamical model and uses model-based reinforcement learning to achieve stable attitude control near the hover state. Within the model predictive control framework, using an accurate data-driven model has been demonstrated to enhance control performance, as shown in previous work~\citep{williams2017information, kabzan2019learning} on racing cars. Similarly, ~\citep{saviolo2022physics} designs MPC based on models learned from real-world data using a physics-inspired Temporal Convolutional Network. Alternatively, rather than learning the full dynamics, a series of works employ machine learning methods in the MPC formulation to learn a robust augmented model that consists of both a first-principle nominal model and a data-driven residual dynamical model. For example, ~\citep{torrente2021data} uses the Gaussian Process to account for aerodynamic effects that arise due to the fast ego-motion of the quadrotor. ~\citep{chee2022knode} proposes KNODE-MPC, which explicitly incorporates the prior physical knowledge (nominal model) into the learning of the augmented model using NeuralODE~\citep{chen2018neural}.

Real-time adaptation to uncertainties is critical for robots operating in dynamic and uncertain environments. Following this direction, online (active) learning~\citep{saviolo2023active} and meta-learning~\citep {richards2021adaptive} techniques are increasingly used in model-based control design. ~\citep{jiahao2023online} extends the KNODE-MPC~\citep{chee2022knode} to an online setting, which recursively constructs a real-time data-augmented dynamical model during deployment. In addition to retraining or learning a new model, one can fine-tune an offline-trained model using real-time data, such as adapting the weights on the last layer of a DNN-represented parametric uncertainty~\citep{joshi2021asynchronous}. A closely related work to our \textit{Proto}-MPC is NeuralFly~\citep{o2022neural}, which uses a DNN basis function to learn the shared representations of various strong wind conditions. NeuralFly explicitly removes task-specificity from the learned DNN through adversarial learning. Consequently, to ensure a stable update of linear coefficients of the basis functions during operation, a Kalman-filter estimation is required to regulate the covariance of the DNN outputs. While effective, it introduces additional estimation and control gain tuning. 
On the other hand, \textit{Proto}-MPC, equipped with an encoder for shared representation and a set of task-specific prototype decoders, not only effectively generalizes across diverse tasks but is also capable of quickly adapting to dynamically changing task conditions.

\subsection{Algorithms}
\begin{algorithm}
\SetKw{KwRequire}{Require:}
\SetKw{KwInput}{Input:}
\SetKw{KwOutput}{Output:}
\KwRequire{ Risk threshold: $R_0$}\;\\
\KwInput{Training dataset: $\mathcal{D}=\{D^{T_k}\}_{k=1:N}$}\;\\
\KwOutput{Encoder $\phi_{\theta}$ and prototype decoder set $\mathbf{W}=\{\mathbf{w}_k\}_{k=1:N}$}\;
\BlankLine
\For{$k=1,2, ..., N$}{
Random initialize $w$ \Comment{Pretrain to ensure achievability (\ref{achi_set})}\;\\
\While{$\mathcal{R}^{T_k}(w, \phi_\theta) > R_0$}{
$w \leftarrow \min_{w\in\mathcal{W}} \mathcal{R}^{T_k}_n(w, \phi_{\theta}) $\;\\
$\theta \leftarrow \theta - \epsilon \nabla_{\theta} \mathcal{R}^{T_k}_n(w, \phi_{\theta})$\;
}}
\BlankLine
\While{not done}{
\For{$k=1,2, ..., N$}{
$\mathbf{w}_k \leftarrow \arg\min_{w\in \mathcal{A}^{T_k}(w)} \mathcal{R}^{T_k}_n(\phi_{\theta}, w)$ \Comment{Compute prototype decoder~\eqref{eq:emp prototype}}
}
random sample $T_i \sim \mathcal{T}$\;\\
$\theta \leftarrow \theta - \epsilon \Big((1-\beta) \nabla_{\theta} \mathcal{R}^{T_i}_n(\mathbf{w}_i, \phi_{\theta}) - \beta \sum_{\mathbf{w'} \in \mathbf{W}\setminus\{\mathbf{w}_i\}} \nabla_{\theta} \mathcal{R}^{T_i}_n(\mathbf{w'}, \phi_{\theta}) \Big)$ \Comment{Meta update~\eqref{regularized_meta_update}}
}
\caption{Training loop for prototype-decoder-based meta-leaning. The empirical risk is computed on the batch dataset that is \textit{uniformly} sampled from the task dataset, i.e. $\mathcal{R}^{T_k}_n(\phi_{\theta}, w) = \frac{1}{n}\sum_{(x_i,y_i)\in D_n^{T_k}} \|y_i - \phi_{\theta}(x_i)w\|^2 $ where $D_n^{T_k} \sim D^{T_k}$}
\label{training}
\end{algorithm}

\begin{algorithm}
\SetKw{Kw}{Require} 
\Kw{: moving horizon data buffer $\mathcal{D}_n$}. The moving horizon data buffer $\mathcal{D}_n(t)$ stores a sequence of real-time data of fixed length $n$, i.e., $\mathcal{D}_n(t)=\{(x_i, y_i) \}_{i=t-n}^{t}$.\;\\
\Kw{: acceptance criterion $D_0$}\;\\
randomly initialize $\mathbf{a}_0, w_0$ \;\\
\For{current time $t=0,1,\dots$}{ 
\eIf{Privileged Information is available}{
$w_t \leftarrow$ \texttt{PI}\big($\mathcal{D}_n(t)$\big) 
}{
$\mathbf{a}_{emp} \leftarrow$ EmpDistribution($\mathcal{D}_n(t)$) \Comment{Compute empirical distribution~\eqref{eq:empirical a}}\;\\
\eIf{$D_{KL}(\mathbf{a}_{emp}\| \mathbf{a}^*) > D_0$}{
Accept $\mathbf{a}_{emp}$ and $\mathbf{a}_t \leftarrow \mathbf{a}_{emp}$ \Comment{Acceptance criterion
}
}{
Reject $\mathbf{a}_{emp}$ and $\mathbf{a}_t \leftarrow \mathbf{a}_{t-1}$}
$w_t\leftarrow \mathbf{a}_t[1]\mathbf{w}_1 + ... + \mathbf{a}_t[N]\mathbf{w}_N$ \Comment{Compute decoder using prototypes}
}
MPC $\leftarrow \hat{f}_{\Delta} = \phi_{\theta}(x)w_t$ \Comment{MPC with adapted residual model}
}
\caption{\textit{Proto}-MPC}
\label{alg:proto MPC}
\end{algorithm}

\newpage
\subsection{Experimental Setup}
\begin{wrapfigure}{r}{0.5\textwidth}
    \vspace{-6mm}
    \centering
    \includegraphics[width = 0.5\columnwidth]{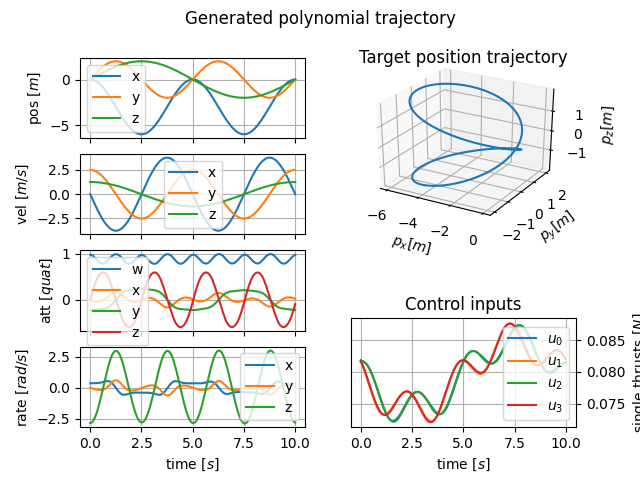}
    \vspace{-6mm} 
    \caption{Reference trajectory for training and data collection.
    }
    \label{fig: reference trajectory}
     \vspace{-5mm} 
\end{wrapfigure}
\textbf{MPC Implementation:} in our implementation, we follow the formulation in equation~\ref{prob: MPC problem} over time horizon $T=1s$ with discretization step $\Delta t = T/N = 1/20s$. We transform the optimal control problem into a nonlinear programming (NLP) via multiple shooting method and solve it using sequential quadratic programming in a real-time
iteration scheme (SQP-RTI)~\citep{diehl2006fast}. The NLP is implemented using acados~\citep{Verschueren2018}.\\

\noindent\textbf{Data Collection}: we consider the polynomial trajectory shown in Figure~\ref{fig: reference trajectory} for data collection, which is obtained using the minimum-snap trajectory generation algorithm ~\citep{mellinger2011minimum}. The data is collected by a quadrotor controlled by a nonlinear MPC with the nominal model $f_{\text{nom}}$. The learning task set is designed for constant side wind in the $x$-direction at speeds of 2, 4, and 6 m/s. For each wind condition, we collected 50 seconds of data for training the EPD model.\\

\noindent\textbf{Training EPD model:} in this experimental setup, the EPD model takes states and controls $[\mathbf{x}, \mathbf{u}] \in \mathbb{R}^{17}$ as its input and outputs the residual lumped forces $\Delta f \in \mathbb{R}^3$. The encoder is a deep neural network of size $[17, 64, 64, 50, 4]$ and the linear decoder is matrix $w \in \mathbb{R}^{4 \times 3}$ with $\sigma(w) < 3.0$. We follow the Algorithm \ref{training} for training the EPD model.\\

\newpage
\subsection{Learning Results}
\begin{figure}[H]
    \floatconts
    {fig: learning results} 
    {\caption{Results for training the EPD model. (a) Ground truth vs. predicted forces on a validation dataset. Note the scale differences. (b) Smoothed batch loss curve with task highlighted.}}
    {
    \subfigure[][c]
    {
    \label{fig:gt vs pred} 
    \includegraphics[width = 1.0\columnwidth]{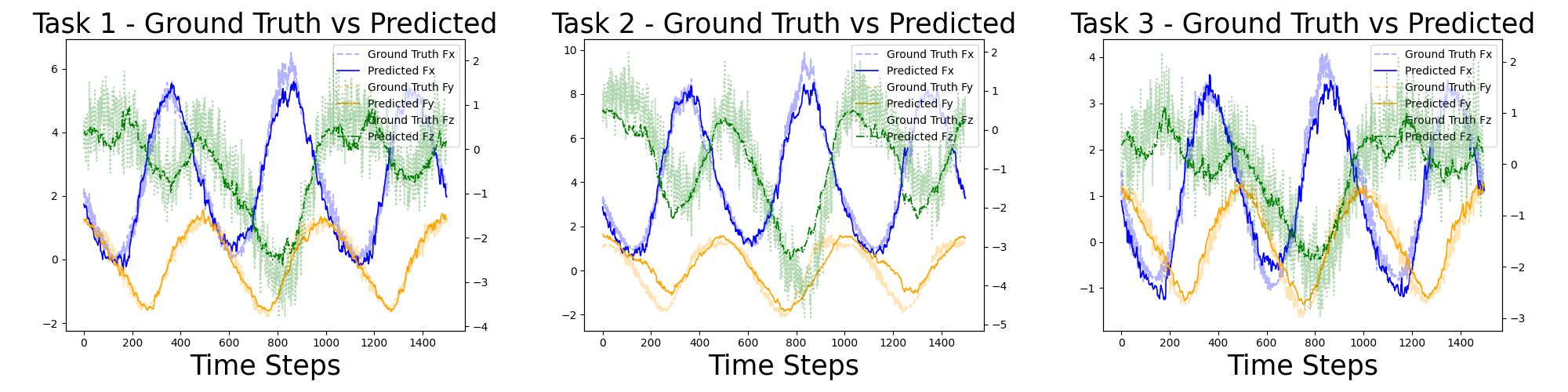} }
    \subfigure[][c]
    { 
    \label{fig:loss curve}
    \includegraphics[width = 0.4\columnwidth]{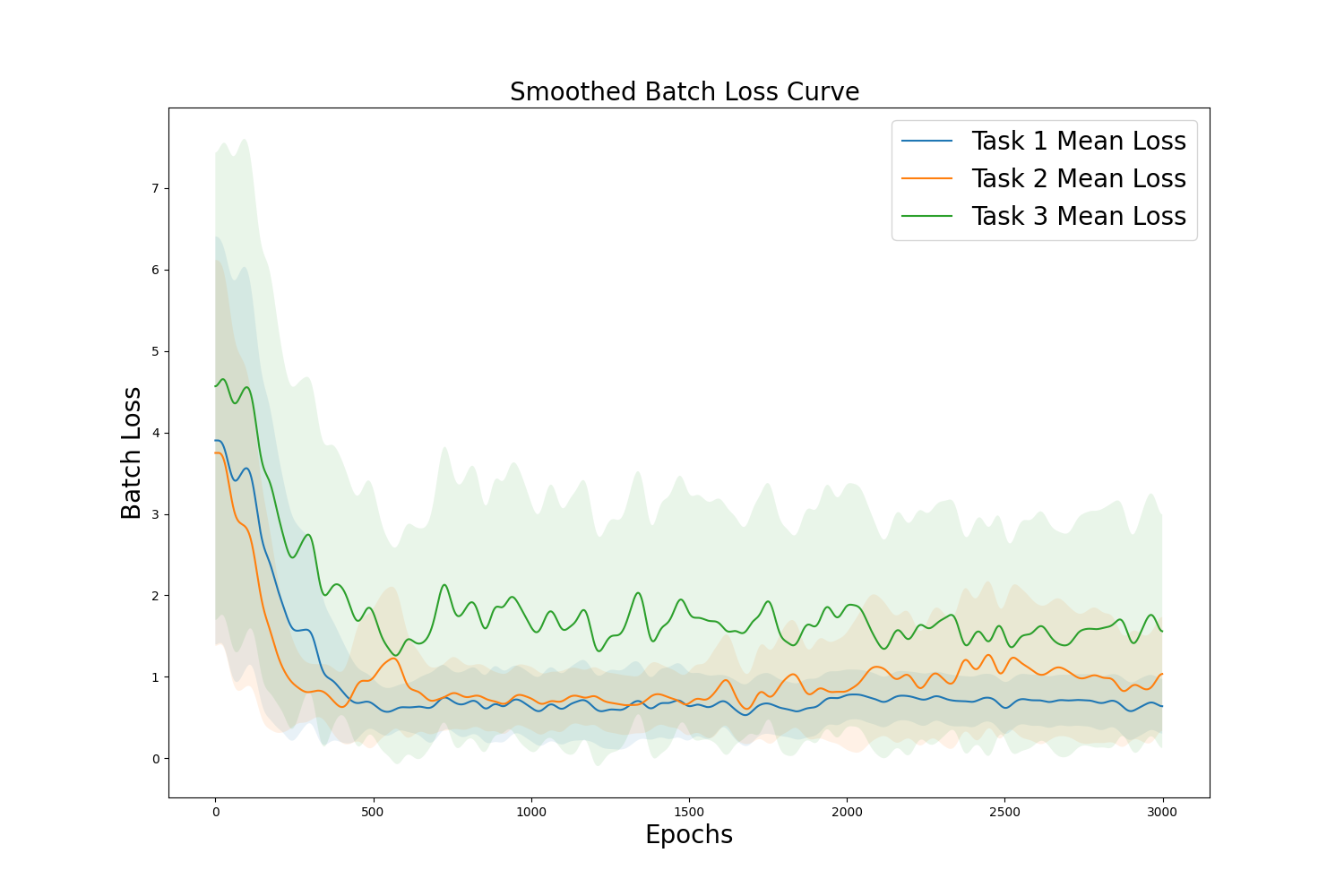}
    }
    }
\end{figure}

Figure~\ref{fig:loss curve} shows the task-specific batch loss curve during training. 
The gradual reduction of the loss indicates that the decoders capture the essential features of their corresponding tasks, while the stable variance band implies a lossy representation of the encoder in a controllable manner, which leaves room for adaptation online. Figure~\ref{fig:gt vs pred} validates the learned network's inference capability on the training task. {The impact of the trade-off parameter $\beta$ to inter-task regularization is discussed in the Appendix.}

\newpage
\subsection{Impact of the trade-off parameter $\beta$ to inter-task regularization}
\begin{figure}[H]
    \centering
    \includegraphics[width=1.0\textwidth]{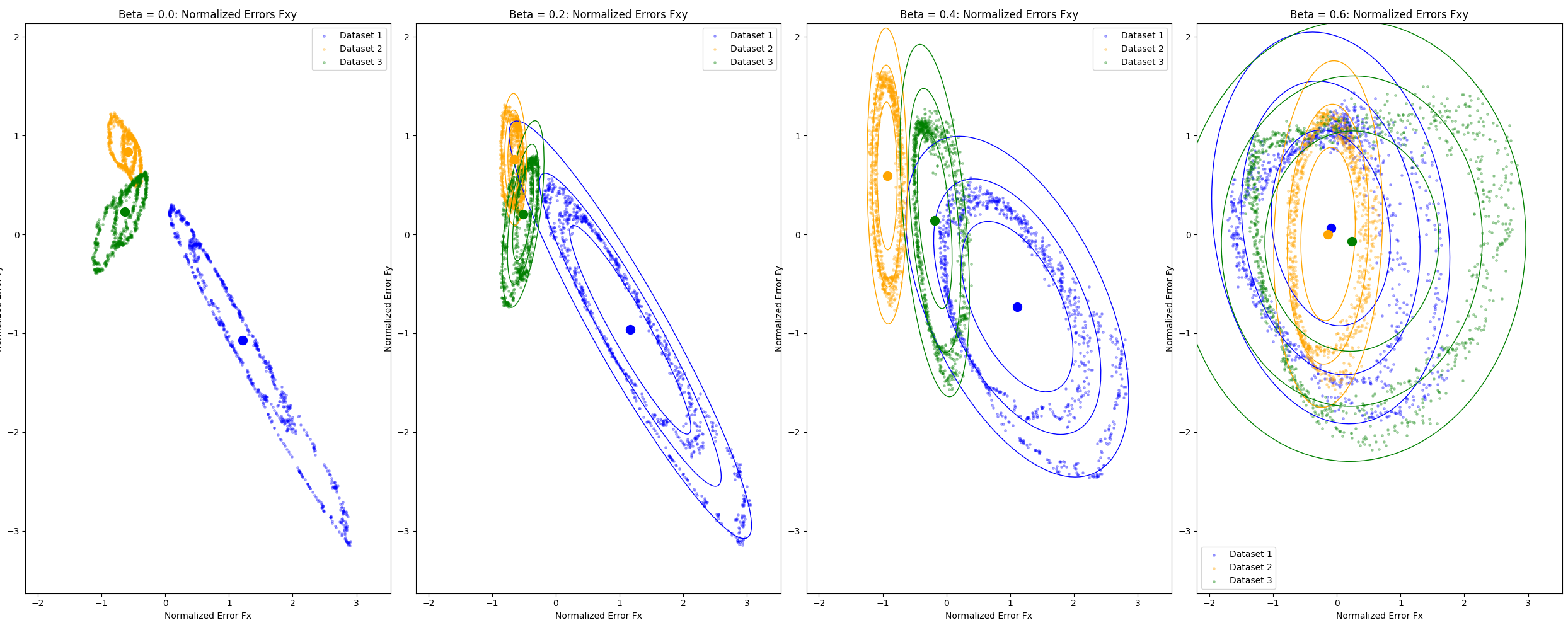}
    \caption{Normalized error distribution with task highlighted: $\beta=$ 0, 0.2, 0.4, 0.6. For better visualization, we only show the task-specific distribution of the normalized risk for the $x$ and $y$ components.}
    \label{fig:clustering}
    \vspace{-0.3 cm}
\end{figure}

Figure~\ref{fig:clustering} illustrates the role of the trade-off parameter $\beta$. The progression across the plots suggests that as $\beta$ increases, the model transitions from task-specific learning to a more regularized task learning. As $\beta=0$, no inter-task regularization is performed, and risks of different tasks show distinctive patterns. \textit{{We highlight the case when $\beta=0.4$}}: the model strikes a desirable balance between different task representations and uniform across-task regularization. The uniformity of the clusters suggests that the encoder is trained to capture the inherent patterns of the residual dynamics. Note that the systematic variation in the $x$-direction of the clusters is aligned with the prior physical knowledge of side winds in the $x$-direction of different intensities.

\subsection{Error distribution for tracking performance subject to spatially varying wind}
\begin{figure}[H]
    \floatconts
    {fig:testing results} 
    {\caption{Error[m] distribution on testing trajectories associated with Fig.~\ref{fig: 3d path testing} subject to spatially varying wind.}}
    {
    \subfigure[][c]
    {
    \label{fig:path 1 mpc rmse} 
    \includegraphics[width = 0.3\columnwidth]{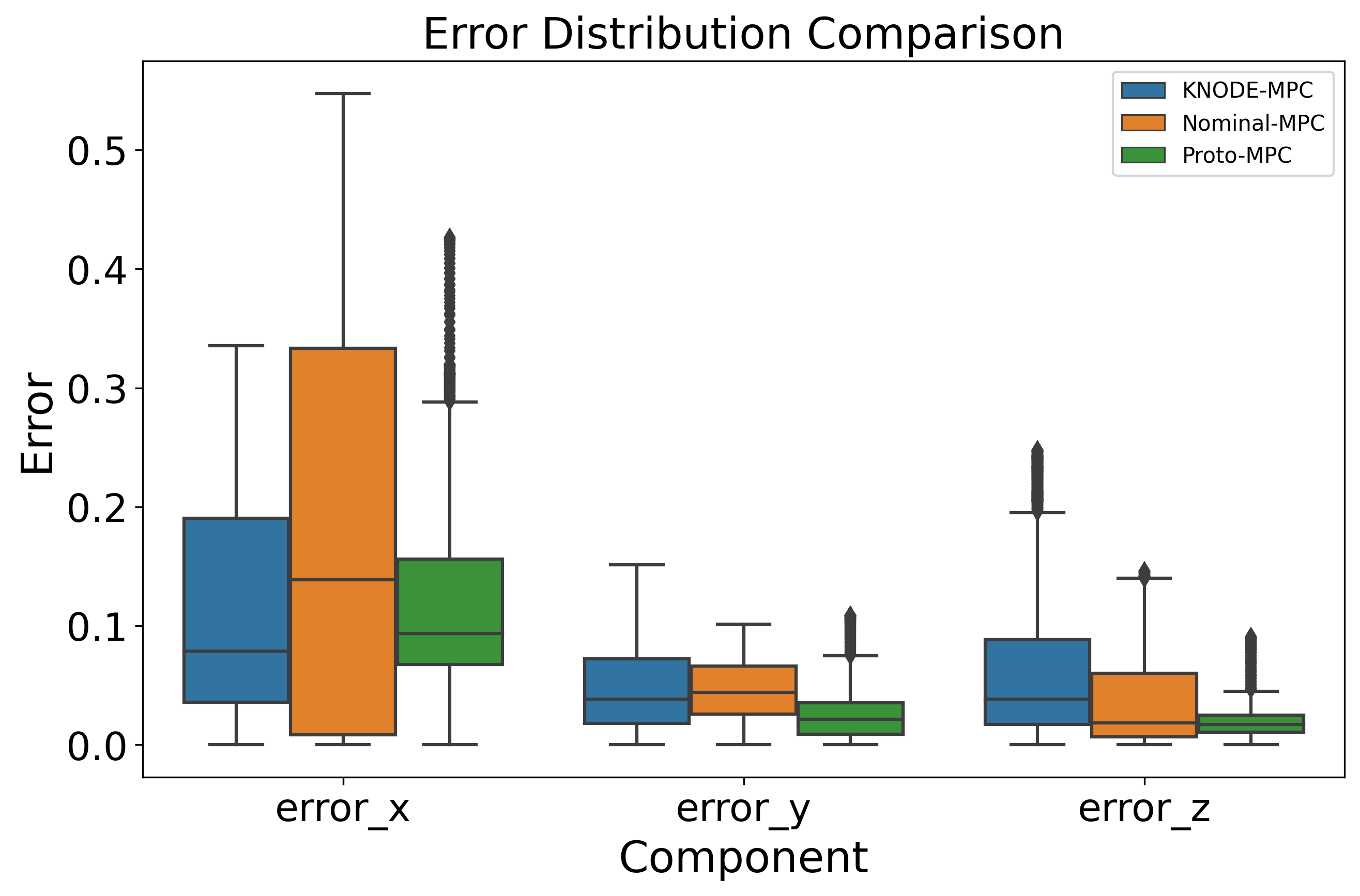} }
    \quad
    \subfigure[][c]
    {
    \label{fig:path 2 mpc rmse} 
    \includegraphics[width = 0.3\columnwidth]{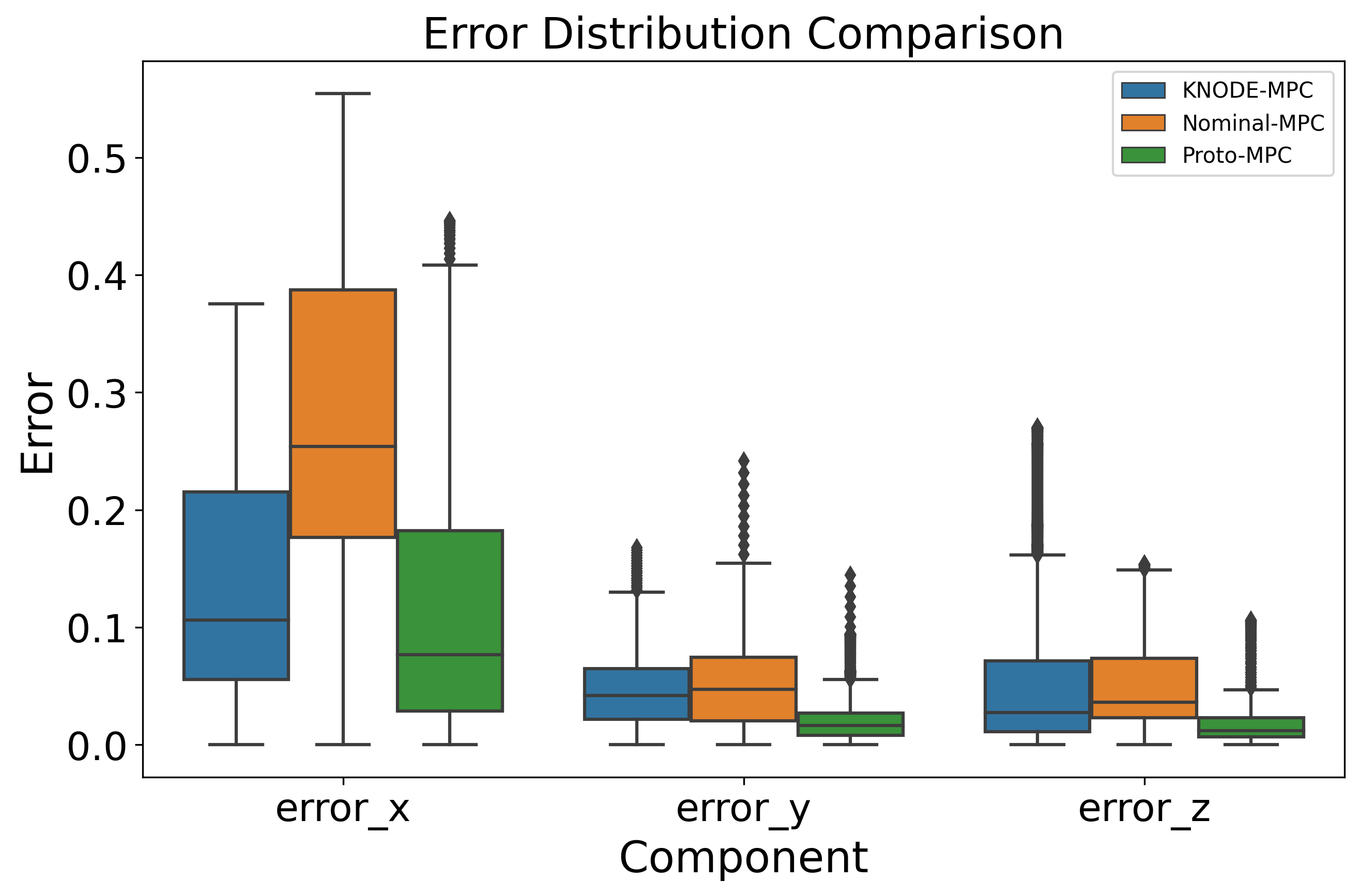} }
    \quad
    \subfigure[][c]
    { 
    \label{fig:path 3 mpc rmse}
    \includegraphics[width = 0.3\columnwidth]{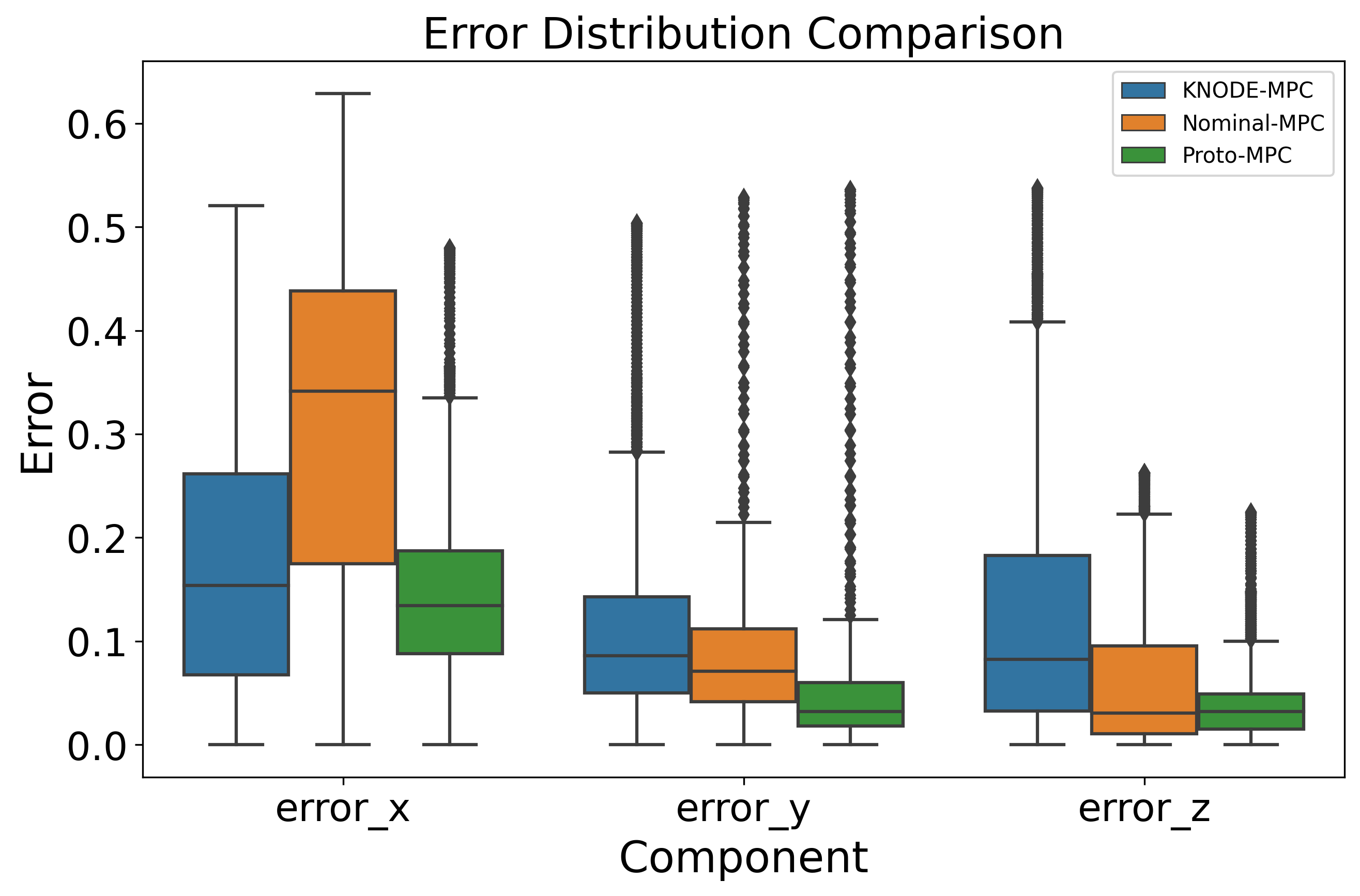}
    }
    }
    \vspace{-0.3 cm}
    \label{fig: 3d path error distribution}
\end{figure}

Figure~\ref{fig: 3d path error distribution} shows the tracking error distribution under the spatially varying wind, supplementing the RMSE results in Table~\ref{tb: rmse testing trajectories}. The box shows the interquartile range of errors from 25th to 75th percentiles. Compared to nominal MPC and KNODE-MPC-online, \textit{Proto}-MPC not only demonstrates reduced mean tracking errors in all components but also shows a more concentrated error distribution, indicating its consistent tracking performance.

\end{document}